\documentclass[11pt,letterpaper]{radixark}

\usepackage{booktabs}
\usepackage{array}
\usepackage{tabularx}
\usepackage{multicol}
\usepackage[flushleft]{threeparttable}

\usepackage{enumitem}

\usepackage{amsmath}
\usepackage{amssymb}
\usepackage{xcolor}
\usepackage{newunicodechar}
\newunicodechar{✅}{\textcolor{green!55!black}{\checkmark}}

\graphicspath{{figures/}}

\setlist[itemize]{leftmargin=1.5em,itemsep=0.25em,topsep=0.35em}
\setlist[enumerate]{leftmargin=1.6em,itemsep=0.25em,topsep=0.35em}

\hypersetup{ pdftitle={Miles: Production-Level Post-Training}, pdfauthor={RadixArk}, pdfsubject={Reinforcement learning post-training systems for large language models}, pdfkeywords={reinforcement learning, post-training, agentic RL, mixture-of-experts, LLM systems} }

\newcommand{\miles}{\mbox{Miles}}
\newcommand{\repourl}{https://github.com/radixark/miles}
\newcommand{\docsurl}{https://miles.radixark.com/docs}
\newcommand{\flag}[1]{\texttt{-{}-#1}}

\newif\ifshowcomments
\showcommentstrue

\begin{document}

\title{Miles v0.1: Production-Level Post-Training}

\author{RadixArk}

\date{\today}
\checkdata[Code]{\protect\url{\repourl}}
\checkdata[Website]{\protect\url{https://miles.radixark.com}}

\abstract{We present \miles{} v0.1, a full-stack, production-ready system for frontier post-training. Building upon the clean design of slime~\citep{slime}, \miles{} designs each stage of the reinforcement-learning (RL) training loop around a single principle: components should be \emph{verified, clean, and customizable}. With accuracy, efficiency, reliability, and scalability as first-class goals, \miles{} aims to make frontier-scale RL accessible to researchers and enterprises alike. This report walks through the system end to end: rollout engines built on SGLang~\citep{sglang}, a trainer with a choice of two backends (NVIDIA Megatron-LM~\citep{megatron} and PyTorch FSDP~\citep{fsdp}), and three weight-synchronization transports for different deployment topologies. Beyond full-parameter RL, \miles{} also supports LoRA RL~\citep{lora}, on-policy distillation, supervised fine-tuning, and true-on-policy rollout-training alignment, and extends the same architecture to diffusion models. We close with an end-to-end case study: fully asynchronous agentic RL on a GLM-5.2 744B-A40B model over terminal-use coding tasks, running on 64 NVIDIA GB300 GPUs with a median step time of 263 seconds over the first 30 measured steps.}

\maketitle

\section{Introduction}
\label{sec:intro}

Post-training turns a pretrained language model into a useful one. At frontier scale, post-training poses substantial challenges for training systems. For example, reinforcement learning (RL) for large language models no longer follows a single generate-then-update loop over short completions. Rollouts now span multiple turns, use tools, and let the model act in an external environment; the models producing the rollouts are often trillion-parameter mixtures of experts (MoE). Such challenges make it difficult to sustain high end-to-end hardware utilization: the system must juggle latency-sensitive rollouts with throughput-oriented training, often introducing bubbles and idle time. Meanwhile, the numerical gap between the rollout engines and the trainer can become large enough to invalidate the objective outright.

To tackle these problems, we present \miles{} v0.1, a full-stack, production-ready system for frontier post-training. \miles{} builds on the clean design of slime~\citep{slime} and centers on one principle: components should be \emph{verified, clean, and customizable}.

The report traces the path of the data through \miles{}, describing the mechanism, the settings that control each part, and what we measured. The report also states the limits of \miles{}: some precision formats remain at an early stage, some weight-transfer paths cover only certain model families, and some measurements come from one configuration rather than many.

\subsection{The Miles RL Loop}
\label{sec:loop}

A reinforcement learning (RL) training job in \miles{} cycles through three stages that generate trajectories, update the policy (the model being optimized), and return the updated weights to generation. A prompt is one task drawn from the dataset; a trajectory is one attempt at that task, and a trajectory group collects trajectories generated from the same prompt:

\begin{enumerate}
\item \textbf{Rollout}: SGLang engines generate trajectories. In agentic RL, where the model acts across multiple turns, each rollout session interacts with its own isolated environment, which executes actions and produces the reward.
\item \textbf{Training}: The trainer uses NVIDIA Megatron-LM~\citep{megatron} or PyTorch FSDP~\citep{fsdp} to consume completed trajectory groups, compute the RL loss, and update the policy.
\item \textbf{Weight update}: After each training step, \miles{} updates the RL policy by synchronizing the new weights with the rollout engines, minimizing interruption to in-flight rollouts.
\end{enumerate}

\begin{figure}[t]
  \centering
  \includegraphics[width=0.85\textwidth]{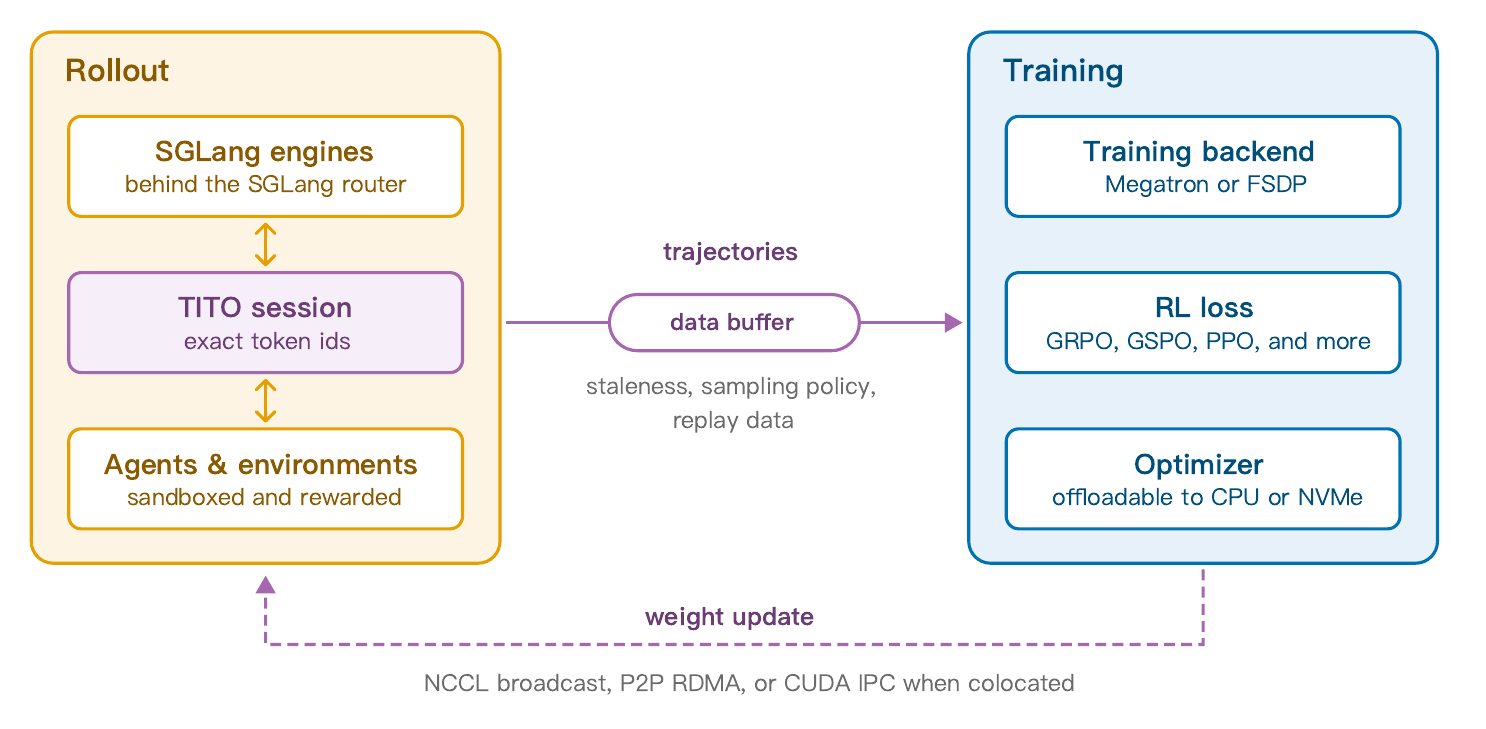}
\caption{The \miles{} RL loop. }
  \label{fig:rl-loop}
\end{figure}

Note that the three stages need not run in lockstep. A schedule that makes the stages take turns idles hardware on both sides: the trainer waits for the slowest trajectory in the batch to return, and the engines then wait for the optimizer to finish. To overlap generation and training, \miles{} offers a fully asynchronous mode in which the engines continue generating while the trainer works, so the two sides can make progress concurrently. Figure~\ref{fig:rl-loop} illustrates the Miles RL loop. 

\subsection{Report Organization}

This report first examines each element of the loop in Section~\ref{sec:loop} and explains how we make \miles{} accurate, efficient, reliable, and scalable. Section~\ref{sec:rollout} covers the rollout stage, including fully asynchronous scheduling, agentic environments, token-in-token-out (TITO) sessions, and rollout routing replay. Section~\ref{sec:training} covers the trainer: low-precision recipes, memory efficiency, the two training backends, and the objective itself. Section~\ref{sec:weight-update} covers weight synchronization.

The report then broadens its scope beyond the core loop. Section~\ref{sec:recipes} covers post-training paradigms beyond core RL, namely LoRA RL, on-policy distillation, and true-on-policy alignment, and Section~\ref{sec:diffusion} extends the architecture to diffusion models. Section~\ref{sec:coverage} sets out verified coverage on the two axes that determine whether a given run is possible at all: day-0 model support (Section~\ref{sec:day0}) and multi-vendor hardware (Section~\ref{sec:hardware}). Section~\ref{sec:code-quality} then states the code-quality principle that organizes the implementation, and Section~\ref{sec:case-study} closes with the end-to-end GLM-5.2 case study on 64 NVIDIA GB300 GPUs.

\section{Rollout}
\label{sec:rollout}

Rollout generation poses two distinct problems in agentic RL: throughput and fidelity. Throughput matters because rollout generation dominates wall-clock time, so request placement and scheduling determine how fast the loop runs. Fidelity matters because the trainer's view of a trajectory can silently diverge from what the policy actually sampled, corrupting the gradient without producing any error. Sections~\ref{sec:fast-rollout} and~\ref{sec:fully-async} address throughput by showing how requests are routed to preserve cache locality and how generation is decoupled from training. Sections~\ref{sec:environments} through~\ref{sec:r3} address fidelity by showing how environments attach to the rollout stack, how token-exact trajectories are recorded across turns, and how expert routing is kept identical between rollout and training.

This section frequently refers to the following four terms, so we define them here. A \emph{prompt} is one task drawn from the dataset. A \emph{trajectory} is one attempt at that task by the current policy. For single-turn work, a trajectory is a single completion; for agentic work, it is an entire multi-turn episode comprising the model's messages, its tool calls, and the environment's replies. A \emph{group} is the set of trajectories generated from the same prompt. The grouping matters because the objectives \miles{} targets, GRPO~\citep{grpo} among them, score a trajectory by comparing it against the other trajectories for the same prompt. Consequently, trajectories sharing a prompt have to be generated and consumed together, so \miles{} schedules, buffers, and discards data one group at a time rather than one trajectory at a time. A \emph{session} is what the serving layer sees, namely the ordered set of requests belonging to a single episode. Note that a session may produce one or more trajectories.

\subsection{Fast Agentic Rollout with SGLang}
\label{sec:fast-rollout}

\miles{} builds on SGLang's fast serving stack and its optimizations for agentic workloads. This section does not intend to detail the optimizations in SGLang; instead, it explains how \miles{} integrates SGLang to turn fast inference into high-throughput agentic RL training.

Request routing is central to this integration because it largely determines throughput for multi-turn rollouts. Each turn of a trajectory reuses nearly all of the context from earlier turns, and the engine that served the previous turn already holds that prefix in its KV cache. If the router sends the next turn to a different engine, that engine must prefill the entire history again, which can drastically reduce rollout throughput as the number of turns grows.

\miles{} therefore fronts its SGLang~\citep{sglang} engine fleet with SGLang's router and configures it to keep all requests from the same multi-turn episode on the same rollout engine. We refer to this cache-preserving binding as \emph{affinity}. Affinity binds a serving-layer \emph{session} (the ordered set of requests for one episode) to the holder of its KV cache for the session's lifetime, so each turn prefills only the newly generated suffix. \miles{} applies affinity at two granularities: session-aware routing binds a session to the engine holding its prefix; and DP-rank-aware routing further narrows that binding to an individual data-parallel rank when DP attention is enabled. In contrast, load-based routing makes a fresh decision for every request and can separate the turns of one session, sacrificing prefix reuse.

In practice, \miles{} uses SGLang's key-based routing mode for affinity: it attaches a stable routing key to every request in a session, and the router consistently maps that key to the same destination (the same engine, or the same data-parallel rank when DP attention is enabled). Load-based routing remains available for single-turn workloads, where requests are independent and prefix reuse does not apply.

\miles{} selects that mode automatically whenever session server is enabled. Session server is the \miles{} component between the agent and the engines that takes ownership of a multi-turn trajectory. The server holds the conversation history, decides how that history becomes tokens, and preserves the exact token IDs produced by the engines, so the trainer later sees precisely what the policy sampled. Section~\ref{sec:tito} describes the session server in full. The server matters here because it tracks session identity and can therefore supply a routing key. A trajectory that reaches the router carrying no key fails outright: \miles{} raises an error rather than routing it by load, because a silent fallback would cost the cache hit rate and leave nothing in the logs to explain why.

Affinity alone can still imbalance the engine fleet. Pinning each trajectory to one engine means that an engine assigned several long trajectories falls behind, while an engine assigned short trajectories runs out of work. The session server therefore strategically decides how a \emph{new} session picks its engine: each new session goes to the engine with the fewest active requests and then stays there for its lifetime. Affinity and least-loaded placement together hold the prefix-cache hit rate at 96\% in the reference run of Section~\ref{sec:case-study}.

\subsection{Fully Asynchronous RL}
\label{sec:fully-async}

Fully asynchronous RL allows rollout generation and training to progress concurrently in \miles{}\footnote{\url{\docsurl/user-guide/fully-async}}. The rollout engines generate trajectories continuously while the trainer consumes whichever trajectories have finished when it needs a batch, so the system doesn't alternate between training and rollout phases. Concurrent execution requires GPU capacity for both stages at once, so the stages use separate GPU pools, a \emph{disaggregated} placement. Note that \miles{} refuses to start a fully asynchronous run when the trainer and rollout engines share GPUs (a \emph{colocated} placement).

The overlap between rollout and training removes the idle time that a slow trajectory creates in a synchronous, turn-taking schedule. When generation and training take turns, a batch completes only after its slowest trajectory finishes, leaving most GPUs idle before that point; the trainer also waits because it needs the complete batch. Long-context and tool-using tasks spend a large part of each batch in this idle state. Under the asynchronous schedule, later batches generate alongside the straggler while the trainer consumes batches that have already completed.

In practice, rollout generation runs continuously; the only point where the rollout engines must pause is when they receive and install a new set of weights from the trainer. The trainer then forms each minibatch by pulling whichever trajectory groups have already finished from the buffer.

\subsubsection{Keeping the Rollout Engines Busy}
\label{sec:submission}

\miles{} keeps the rollout engines busy by replenishing rollout generation capacity as trajectories finish, which aims to minimize trainer wait time. A background worker maintains a number of trajectories in generation and starts a new trajectory based on the replacement rule, which determines how closely generation stays to the limit. The worker can wait for a whole group to finish, or it can reclaim each trajectory's place as soon as that trajectory finishes. \miles{} lets a run choose between the two rules.

\begin{itemize}
\item \textbf{Group granularity} waits until every trajectory in a group finishes before starting a replacement. A single slow trajectory therefore keeps the group's entire share of the limit occupied, so the rollout engines run below the limit until that trajectory ends.
\item \textbf{Sample granularity}, the default under fully asynchronous rollout, lets each finished trajectory free its own place immediately and starts a replacement group once enough places are free. The number of trajectories generating at once therefore stays near the limit even when trajectory lengths differ by an order of magnitude.
\end{itemize}

\subsubsection{Data Buffer Between Generation and Training}
\label{sec:data-buffer}

\miles{} places every finished rollout group in a bounded data buffer before the trainer consumes it. The buffer decouples the rollout engines from the trainer, absorbing differences in their rates so neither stage must match the other's speed. The buffer also provides the single decision point at which \miles{} determines whether a finished group is worth training on, as described below. The implementation is deliberately replaceable: it exposes only three operations: put a group in, take a batch out, and report its metrics. A user who wants different data to reach the trainer can implement their own selector with the same three operations, and point \miles{} at it by its import path. Nothing else in the loop needs to change.

The buffer can discard a group under one of the three conditions, and it checks each condition at a different point (Table~\ref{tab:buffer-policy}). The first two conditions are fixed properties of the group, so the buffer checks them as soon as the group arrives. The third condition, \emph{Staleness}, instead compares the current weights with the oldest weight version under which any turn in the group was generated. A group may therefore arrive stale when its turns span several weight updates during generation. As the group waits in the buffer, the trainer continues updating the weights, so the group can become even staler. Consequently, the buffer only checks staleness when the group is collected by the trainer.

\begin{table}[t]
  \centering
\small
  \begin{tabularx}{\textwidth}{@{}l >{\raggedright\arraybackslash}X l l@{}}
    \toprule
    \textbf{Why the group is dropped} & \textbf{Typical case} & \textbf{Checked} & \textbf{Prompts afterwards} \\
    \midrule
    Generation gave up on it & An agentic episode exceeded its collection timeout, so the group never completed & On arrival & Retried or discarded \\
    \addlinespace
    A user filter rejects it & Every attempt at the prompt received the same reward, so the group carries no advantage signal & On arrival & Discarded \\
    \addlinespace
    Its weights are too old  & The trainer advanced past the staleness limit while the group waited in the buffer & On the way out & Retried or discarded \\
    \bottomrule
  \end{tabularx}
\caption{The three reasons the buffer drops a finished group. The first two are properties of the group, so they are checked as soon as it arrives; staleness depends on how long the group waited, so it is checked only when the trainer collects it. The run supplies the filter and the staleness limit and chooses whether dropped prompts are retried or discarded, except that filter-rejected groups are always discarded because they carry no gradient signal.}
  \label{tab:buffer-policy}
\end{table}

The buffer has a bounded capacity, which is set as a multiple of the training batch size. Once the buffer is full, adding a group blocks until the trainer consumes one. A group goes unconsumed when the buffer drops it, which happens in the three cases of Table~\ref{tab:buffer-policy}: generation gave up on the group, a filter rejected it on arrival, or its weights aged past the staleness limit while it waited. \miles{} then either discards the group or returns its prompts to the data source, so fresh trajectories can be generated for them later.

Staleness warrants a precise definition here, because tokens in one group may use different weight versions. \miles{} defines a group's staleness as the current trainer weight version minus the \emph{oldest} weight version appearing anywhere in the group. Note that this definition is deliberately pessimistic. A group is therefore never treated as fresher than its oldest token.

\subsubsection{Observability}
\label{sec:async-metrics}

In fully asynchronous RL, monitoring the data buffer in Section~\ref{sec:data-buffer} is necessary because rollout and training advance at independent rates. The two stages can drift apart until one outruns the other, wasting hardware silently without crashing or raising an error. When the rollout engines finish groups faster than the trainer consumes them, groups pile up and age in the buffer; some eventually exceed the staleness limit and are discarded, wasting the GPU time that produced them. On the other hand, when the trainer consumes groups faster than the rollout engines produce them, the buffer empties and every training step stalls waiting for a batch, exactly the wait that Section~\ref{sec:submission} set out to remove. The buffer sits between the two stages, so its state distinguishes a growing backlog from an empty queue: the buffer determines which groups the trainer sees and how stale those groups are, while \miles{} reports the quantities in Table~\ref{tab:async-metrics} on every training step.

\begin{table}[t]
  \centering
\small
  \begin{tabular}{@{}l l@{}}
    \toprule
    \textbf{Metric} & \textbf{Reports} \\
    \midrule
    \texttt{queue\_size}               & Groups waiting in the buffer when the step collected its batch \\
    \texttt{avg\_staleness}            & Mean staleness of the groups this step drew from the buffer \\
    \texttt{max\_staleness}            & Highest staleness among the groups this step drew \\
    \texttt{buffer\_avg\_staleness}    & Mean staleness of the groups still waiting \\
    \texttt{buffer\_max\_staleness}    & Highest staleness among the groups still waiting \\
    \texttt{aborted\_groups\_filtered} & Groups dropped on arrival because generation gave up \\
    \texttt{stale\_groups\_filtered}   & Groups dropped at collection for exceeding the staleness limit \\
    \bottomrule
  \end{tabular}
\caption{Buffer metrics reported on every training step, under the \texttt{rollout/fully\_async/} prefix. Staleness appears twice: once for the groups a step drew, and once for the groups still waiting.}
  \label{tab:async-metrics}
\end{table}

Queue size is the quickest signal for identifying which stage limits progress. A queue size pinned at zero means the rollout engines cannot keep pace, so rollout capacity must grow in order to reach maximal efficiency. By contrast, a queue size pinned at capacity means the trainer is the constraint, and the staleness of the waiting groups climbs alongside the queue. Between those extremes, a rising count of groups discarded for staleness indicates that groups are aging out faster than the trainer consumes them.

\subsubsection{Asynchronous Evaluation}

Evaluation competes with training-data generation under a fully asynchronous schedule, so its cost depends on which engines run it. Under a synchronous schedule, the rollout engines sit idle throughout the training phase, making evaluation in that window nearly free. Under a fully asynchronous schedule, the rollout engines continuously generate training data, so evaluation on those engines displaces generation. \miles{} therefore offers three asynchronous evaluation modes, detailed in Table~\ref{tab:eval-modes}.

\begin{table}[t]
  \centering
\small
  \begin{tabular}{@{}l l l l@{}}
    \toprule
    \textbf{Mode} & \textbf{Selected by} & \textbf{Weight source} & \textbf{Effect on training} \\
    \midrule
    Shared engines  & (default)              & Live rollout fleet  & Rollout production pauses \\
    Dedicated fleet & Reserving evaluation GPUs & Checkpoint snapshot & Export may pause; eval is async \\
    External        & A checkpoint backend      & Checkpoint directory & Export may pause; eval is async \\
    \bottomrule
  \end{tabular}
\caption{The three evaluation modes under fully asynchronous rollout. Shared engines measure whichever weights the fleet most recently received, and new generation stops while they do so, although in-flight requests finish. The snapshot-based modes measure the exact weights in the snapshot they receive; an external backend sees only a checkpoint directory, so it can be any service, with or without SGLang.}
  \label{tab:eval-modes}
\end{table}

Snapshot-based modes keep evaluation off the critical path after exporting the snapshot. However, exporting a fresh snapshot imposes a pause because it is a collective operation across the training actors, so the trainer has to wait for the export.  Reusing a periodically saved checkpoint can avoid such pause. Once the snapshot exists, the trainer hands the evaluation off and continues training without waiting for the evaluation results.

A returned evaluation score must identify the policy version that produced it. \miles{} enforces and keeps track of such correspondence rather than assuming it. When an evaluation finishes several steps late, \miles{} records its score against the step whose weights it measured and reports the delay separately, so users can see how late the score arrived. The dedicated evaluation fleet also verifies weight delivery before evaluating: it loads the snapshot onto every engine and confirms that each one reports the expected version. Note that weight verification matters here because the router spreads requests across the entire fleet: an engine that still holds older weights can return a score silently mixed across two weight versions. \miles{} checks two conditions that confirm whether a score represents the correct weight version: the weight version, averaged across every sample the evaluation produced, equals the step against which the score was recorded; and the fraction of requests served by a mixed set of versions is zero.

\begin{figure}[t]
  \centering
  \includegraphics[width=0.85\textwidth]{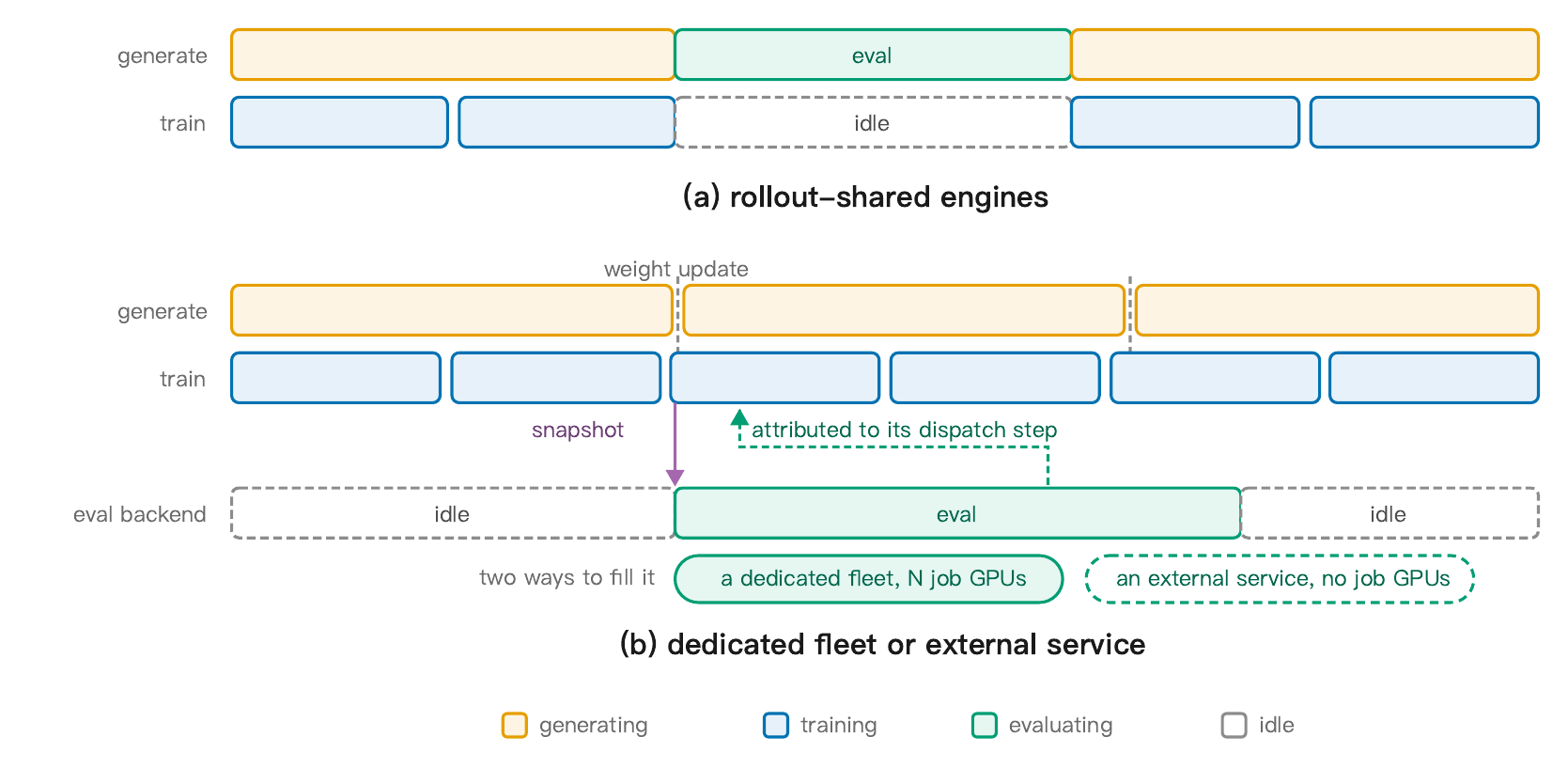}
\caption{The three evaluation modes on the training timeline. Sharing the rollout engines halts generation while the evaluation runs. The two snapshot-based modes run alongside training, although exporting a fresh snapshot or having too many evaluations already outstanding can still pause the trainer.}
  \label{fig:eval-modes}
\end{figure}

Evaluation failures do not stop training, even when snapshot export or verification fails. When an evaluation fails, \miles{} records it as skipped and logs the reason: an export failure, a missing snapshot, too many evaluations already outstanding, or an error inside a user-supplied backend. The timeline in Figure~\ref{fig:eval-modes} places the three modes alongside training.

\subsection{Agentic Environments}
\label{sec:environments}

Agentic RL needs an integration boundary that accommodates environments with different scopes of control. Each trajectory comes from an environment; a coding-agent environment, for example, provides a sandbox per task where the model runs commands, edits files, and receives a final grade from a test suite. Some environments only own the episode loop, whereas others also take charge of managing batching, rewards, and token recording. Consequently, one common agentic interface would either constrain the first kind of framework or duplicate the responsibilities that the second kind already owns.

To accommodate different scopes, \miles{} organizes integration as three nested plug-in layers, and each connector replaces exactly one layer, as is shown in Table~\ref{tab:plugin-layers}.

\begin{table}[t]
  \centering
\small
  \begin{tabular}{@{}l c c c@{}}
    \toprule
    & \textbf{Agent fn.} & \textbf{Generate fn.} & \textbf{Rollout fn.} \\
    & \textit{(innermost)} & & \textit{(outermost)} \\
    \midrule

    Agent--environment loop                     & \checkmark & \checkmark & \checkmark \\
    Trajectory and token recording              & $\circ$    & \checkmark & \checkmark \\
    Group rewards        & $\circ$    & $\circ$    & \checkmark \\
    Data source (prompts, task set)             & $\circ$    & $\circ$    & \checkmark \\
    Batch orchestration (grouping, filtering)   & $\circ$    & $\circ$    & \checkmark \\
    Model, engines, weight updates, advantages, optimizer   & $\circ$    & $\circ$    & $\circ$ \\
    \bottomrule
  \end{tabular}
\caption{Three nested rollout plug-in layers. The agent function is innermost, and each column to its right wraps the one before it; a \checkmark~means the external framework assumes that responsibility, and $\circ$ means \miles{} retains it. Group rewards are scores that need the whole group at once, such as ranking the trajectories against one another, as distinct from GRPO's group-relative centering, which \miles{} always applies itself.}
  \label{tab:plugin-layers}
\end{table}

Miles ships connectors that occupy specific layers in this structure. Harbor~\citep{harbor}, NeMo~Gym~\citep{nemogym}, and OpenEnv~\citep{openenv} connectors attach at the agent function, so the session server described in Section~\ref{sec:tito} records their tokens. By contrast, HUD~\citep{hud}, Strands Agents~\citep{strands}, and $\tau$-bench~\citep{taubench} connectors attach at the generate function and take over token recording as well. The Prime Intellect Verifiers~\citep{verifiers} connector replaces the rollout function outright: it brings its own taskset, groups the episodes, and computes per-rollout and group rewards before returning completed traces. If a user wants to supply their own environment, they can attach through any of the same three layers.

The choice of sandbox backend is independent of the connector layers: a sandbox provider supplies task containers \emph{inside} a connector rather than occupying a layer of its own. \miles{} exercises AgentENV~\citep{agentenv}, Daytona~\citep{daytona}, E2B~\citep{e2b}, and Modal~\citep{modal}, across Harbor, HUD, NeMo~Gym, and OpenEnv. Sandbox lifetime is likewise a recipe choice. For example, the terminal-bench recipes build one sandbox per episode from the task image and delete it afterwards, whereas \miles{} can also point every episode at a single long-running environment server, reusing that server across episodes rather than rebuilding it.

We note two limitations. First, the connectors and sandbox integrations named here are experimental and continue to evolve. Second, the session server does not record screenshots yet, so computer-use connectors record their own trajectories at the generate-function layer; once it does, they can attach at the agent function instead.

\subsection{Token-In-Token-Out (TITO)}
\label{sec:tito}

Multi-turn agentic RL requires the trainer to see exactly the tokens the policy sampled. The rollout engines produce one token sequence, and the trainer later reconstructs a sequence from messages; training is faithful only when those sequences match. A conventional multi-turn pipeline offers no such guarantee because model outputs pass through message parsing, tool execution, and chat-template rendering before the next turn begins. Each stage may change tokenization, prune historical reasoning, or reserialize tool calls. A single change silently breaks the correspondence between sampled tokens and the trainer's log-probabilities, so the update is taken with respect to a trajectory that never occurred. This discrepancy appears as train-rollout mismatch: the serving and training sides assign different probabilities to what is nominally the same trajectory. Consequently, the importance ratio drifts away from one (Section~\ref{sec:objective}), and, if left uncorrected, the mismatch degrades the policy it is supposed to improve.

The TITO session server\footnote{\url{\docsurl/user-guide/agentic-rollout}}~\citep{miles_tito} closes that gap by letting the server, rather than the harness, control tokenization. The agent exchanges ordinary messages and sends its full history on every turn, but the server decides how those messages become tokens. On the first turn, the server renders the selected template into token IDs. After each successful completion, the server checkpoints those prompt IDs together with the output token IDs, log-probabilities, and routed experts returned by SGLang. On later turns, the server reuses the deepest applicable checkpoint and tokenizes only the appended suffix. Any token fields that an agent supplies itself are overridden.

\begin{figure}[t]
  \centering
  \includegraphics[width=0.85\textwidth]{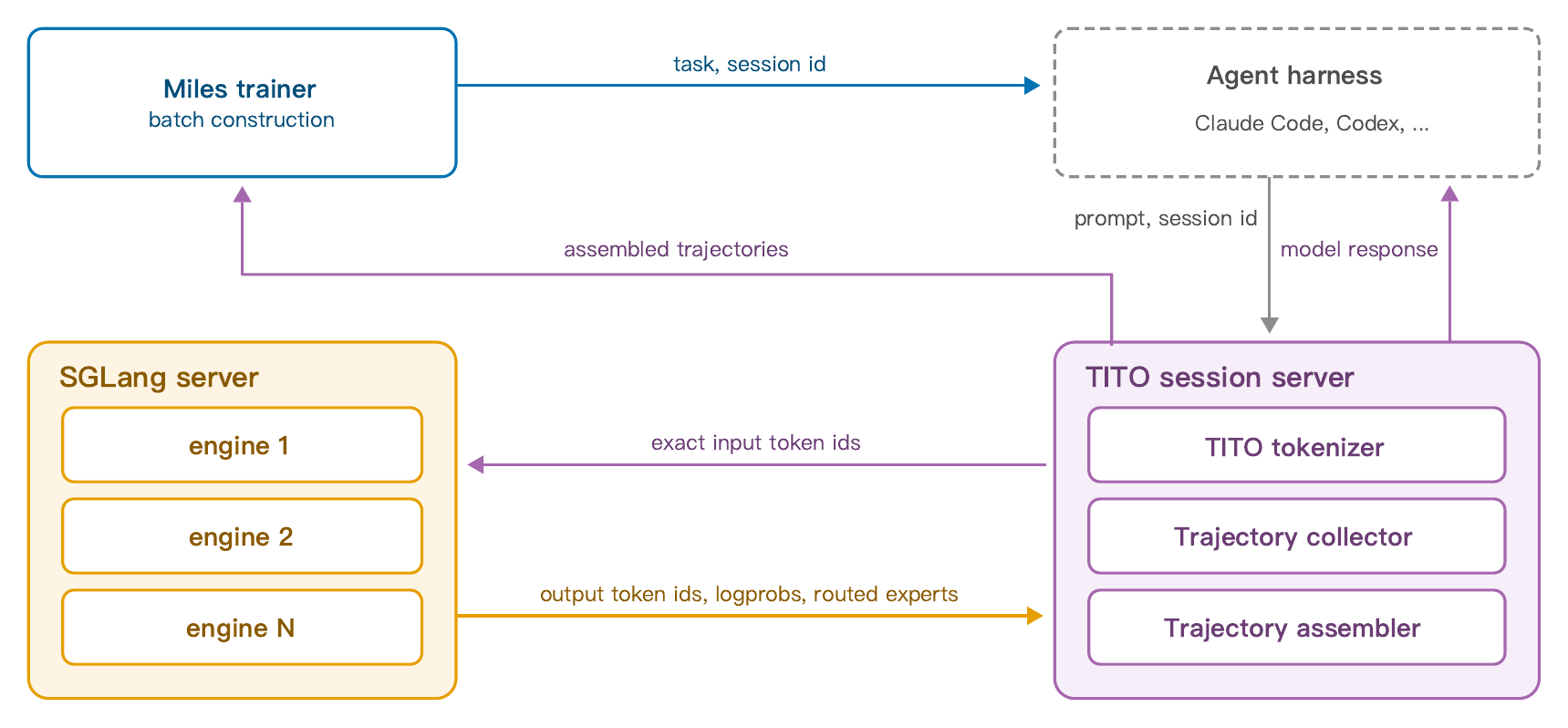}
\caption{Token-in, token-out. The session server preserves the exact token IDs produced by the engines, so the trainer sees the model-generated tokens even when the harness is a black box.}
  \label{fig:tito}
\end{figure}

Server-owned tokenization lets \miles{} assemble the whole trajectory into one contiguous training sequence. The sequence preserves the rollout log-probabilities while loss-masking the tokens the model did not generate. Token fidelity is therefore a precondition for three mechanisms described later: rollout routing replay (Section~\ref{sec:r3}), on-policy distillation (Section~\ref{sec:opd}), and true-on-policy alignment (Section~\ref{sec:zero-kl}). Figure~\ref{fig:tito} shows the assembly that makes all three possible.

\subsubsection{Linear and Branching Sessions}
\label{sec:sessions}

\miles{} supports two rules for extending a session's stored history, linear and branching. The rule a run selects determines how a request can extend a session and how many training trajectories a session can produce. The session server from Section~\ref{sec:tito} owns the stored history and records the token IDs at every completed turn as a checkpoint. When a new request arrives, the server matches its message path against those checkpoints and attaches the request at the deepest applicable checkpoint.

\begin{itemize}
  \item \textbf{Linear.} The server permits each request to extend only the tail of the stored message history. An agent may retry its most recent turn by rolling back one assistant checkpoint, but the server rejects requests that diverge earlier or roll back farther. Consequently, a linear session yields exactly one training sequence.
  \item \textbf{Branching.} The server retains the stored history as an append-only tree. The server attaches each request to the deepest checkpoint whose message path is a prefix of the request; any unmatched remainder opens a new branch, and the server never deletes a branch. Each leaf selected by the server yields one training sequence, so a session can produce multiple trajectories. A branch whose last generation stopped at the length limit cannot be extended.
\end{itemize}

Branching supports harnesses that fork or compact their context while a task runs, e.g.~Claude Code. Note that a harness that reshapes its own history doesn't know in advance how many trajectories a session will produce. As a result, such harnesses cannot be accommodated by the linear rule.

\subsubsection{When Harnesses Do Not Replay Verbatim}

Replay comparison determines whether a session can reuse its stored token prefix, and it is also where token exactness most often fails. Before extending a session under either rule (Section~\ref{sec:sessions}), the session server compares each replayed message with the message already stored at that position. Some harnesses do not replay the model's own messages verbatim: a harness may reserialize tool-call arguments, substitute an empty JSON object for absent arguments, or omit reasoning content on the following request. Under exact comparison, every one of those differences is treated as divergence: a linear session rolls back or rejects the request, whereas a branching session forks a redundant lineage. In either situation, a stored prefix, which should remain valid, is discarded.

To accommodate multiple comparison policies, \miles{} makes the comparison configurable. Three built-in settings range from strictest to loosest, and a user may also supply a comparison policy of their own. The strictest setting is the safe default: it compares only the fields a chat template actually reads, namely the role, the content, the reasoning content, and the tool calls, ignoring fields that no template consumes. This prevents a client library's own additions to a message from being treated as divergence. The next-looser setting accepts tool-call arguments that carry the same JSON content but were serialized differently, while still requiring call identifiers, function names, and ordering to agree. A harness that deviates only in how it serializes those arguments therefore needs nothing looser than this setting.

Looser settings can merge genuinely different histories because they compare fewer fields. The loosest built-in setting illustrates the risk: it compares only the role and the visible text of a message, ignoring tool calls entirely. Two assistant turns then count as the same message whenever their visible text agrees, even when they invoked different tools or the same tool with different arguments. For a tool-calling turn, the visible text is often empty. The stored prefix then prevails, so the trainer keeps a history showing a call the agent never made. Because \miles{} does not reconcile tool-call identifiers when two histories collapse this way, the mismatch remains silent.

Every matcher setting answers one question only: whether a replayed message counts as the message already stored at that position. When the matcher treats the two messages as the same, the stored token snapshot remains authoritative, and the session server tokenizes only the newly appended suffix.

\subsubsection{Verification and Current Limits}

Correct tokenization depends on the chat template and tool call parsers a checkpoint expects, so \miles{} verifies that combination before supporting the checkpoint. The session server described in Section~\ref{sec:tito} uses a registered \emph{model family}, a set of checkpoints that share one chat template and one pair of reasoning and tool-call parsers. \miles{} never detects a family from a checkpoint. A user names the family explicitly when launching the run, and that name resolves to a registration that a maintainer has already checked. Registrations span the Qwen3, GLM, Nemotron, Kimi, MiniMax, DeepSeek, and Inkling lines. A checkpoint outside those lines falls back to a generic handler, and \miles{} warns that its incremental tokenization may diverge from the template's canonical output.

To support a new model for TITO, the user can register the model in \miles{}. Two checks guard each registration. The first check runs on CPU and confirms that rendered token sequences remain append-only from turn to turn, which is the invariant on which the design rests. The second check runs on GPU against a live model and confirms that the invariant survives real inference, where stop tokens and tool-call parsing come into play. It is worth mentioning that the CPU check alone cannot guarantee correctness, because a template can remain append-only in isolation, yet break when a real parser consumes the model's output.

The session server does not yet carry image or video inputs, so \miles{} cannot run vision-language models through it. Consequently, a vision-language model instead drives the SGLang rollout engine directly, through its lower-level token-in, token-out interface.

\subsection{Efficient Rollout Routing Replay (R3)}
\label{sec:r3}

In a mixture-of-experts (MoE) model, rollout and training can send the same token to different experts: for each token, the router scores the experts through a learned projection and keeps the top $k$, and small numerical differences between training and rollout can flip the winning experts independently at every layer and for every token. This means that giving the trainer the exact tokens sampled by the policy (Section~\ref{sec:tito}) still may not reproduce the token-level log-probabilities.

To understand the issue caused by mismatched expert routing, let us consider the following example. Suppose rollout selects experts $\{2, 7\}$ for a token. If the trainer uses different kernels at a different precision with the same weights, its router can select $\{2, 8\}$. The update then reaches an expert that never contributed to the sampled token, while the contributing expert receives no update. Across layers, tens of thousands of tokens per sequence, and thousands of training steps, the mismatch drives the updated policy away from the policy that generated the data. Because the next batch comes from the policy produced by the preceding update, the resulting drift feeds back into training and compounds over a longer run. It is reported in \citet{r3} that such routing discrepancy destabilizes RL dramatically in MoE models and can end in catastrophic training collapse.

R3 is a technique that mitigates this issue by treating each token's expert assignments as part of the rollout data, allowing training to replay the rollout assignments exactly rather than recompute them. A user enables R3 in \miles{} through \flag{use-rollout-routing-replay}, which allows SGLang to return the routed experts alongside the generated tokens. During the forward pass, the trainer replays those assignments, which sends each token to exactly the same experts that processed it during rollout. Because the session server already records routed experts together with token IDs and log-probabilities (Section~\ref{sec:tito}), replay covers entire multi-turn episodes rather than only a single completion.

It is worth mentioning that recorded assignments are cheap to replay but expensive to carry. Each routing tensor holds $(\text{tokens} - 1) \times \text{layers} \times k$ 32-bit integers. For a 32K-token sequence over 60 layers at $k = 8$, that tensor occupies roughly 60\,MB per trajectory, and the payload must remain in memory and move alongside the trajectory. Because the cost grows with sequence length and agentic RL produces long sequences, the overhead can become substantial. Section~\ref{sec:memory} describes the memory headroom that makes such cost affordable at scale.

Note that R3 is not enabled everywhere: \miles{} makes it a per-recipe choice. A dense model has no expert routing to replay, so R3 does nothing for it. In asynchronous RL, since there are many other factors contributing to train-rollout mismatch, enabling R3 might also have only limited effects. Several shipped MoE recipes in \miles{} enable R3, whereas the GLM-5.2 reference run of Section~\ref{sec:case-study} leaves it off.

\section{Training}
\label{sec:training}

The trainer in RL turns rollout data into weight updates. Four trainer properties underpin a frontier-scale run: the numerical precision in which it operates (Section~\ref{sec:low-precision}), whether the optimizer state fits in memory (Section~\ref{sec:memory}), the backend that owns the model on the GPU (Section~\ref{sec:backends}), and the objective it optimizes (Section~\ref{sec:objective}). This section examines those properties in that order.

\subsection{Low-Precision Training}
\label{sec:low-precision}

Lower-precision number formats make matrix multiplication faster. A GPU's tensor cores roughly double their peak rate each time the precision halves, and successive NVIDIA generations have put the low precisions within reach, with 8-bit floating point on Hopper and 4-bit on Blackwell. Proper quantization in RL is, however, not straightforward. Quantizing rollout alone, or quantizing rollout and training differently, means that the two sides compute differently from the same weights, and the pipeline accumulates that disagreement layer by layer, potentially leading to catastrophic train-rollout mismatch.

Therefore, low-precision training requires a shared contract between rollout and training: both stages must quantize the same weights in the same way. \miles{} enforces that contract with a single low-precision path: rollout and training run the same quantization logic on the forward pass. Three quantized formats have an end-to-end recipe today, listed in Table~\ref{tab:precision-formats}. Two further options sit alongside them: quantization-aware training in INT4, and a ``BF16 train, FP8 serve'' mode that is easier to stand up when a new architecture first comes online. Separate reports cover the FP8~\citep{miles_fp8} and INT4~\citep{miles_int4} recipes in depth.

\begin{table}[t]
  \centering
\small
  \begin{tabular}{@{}l l l >{\raggedright\arraybackslash}p{3.3cm} l@{}}
    \toprule
    \textbf{Format} & \textbf{Block} & \textbf{Scales} & \textbf{Models tested} & \textbf{Maturity} \\
    \midrule
    BF16          & --                 & --        & All                          & Baseline \\
    FP8 blockwise & $128 \times 128$    & FP32       & Qwen3-4B, Qwen3-30B-A3B, DeepSeek-V4 & Generally available \\
    MXFP8         & $1 \times 32$       & UE8M0      & Qwen3-30B-A3B, DeepSeek-V3.2 & Beta \\
    NVFP4 (E2M1)  & $1 \times 16$       & E4M3, FP32 & Qwen3-30B-A3B                & Beta \\
    \bottomrule
  \end{tabular}
\caption{Low-precision formats with an end-to-end \miles{} recipe, against the BF16 baseline. NVFP4 nests an E4M3 scale per block inside one FP32 scale per tensor. FP8 blockwise runs on NVIDIA Hopper and Blackwell and on AMD MI350X and MI355X; MXFP8 and NVFP4 need Blackwell; A100 has no FP8 arithmetic and runs BF16 only.}
  \label{tab:precision-formats}
\end{table}

Note that rollout and training cannot mix those formats freely. The general rule permits two arrangements: the two sides run the same format, or the trainer stays in BF16 while rollout quantizes. A special case is NVFP4, which requires every stage that touches the weights to quantize them.  Since a BF16 trainer by definition quantizes nothing, pairing NVFP4 rollout with a BF16 trainer is unsupported.

\miles{} builds the MXFP8 and NVFP4 recipes~\citep{miles_mxfp8_nvfp4} as end-to-end \emph{precision contracts}. Each contract checks four stages to ensure that the same quantization agreement is honored: checkpoint conversion, the trainer's forward pass, SGLang rollout, and the live weight export that converts the trainer's weights into the rollout engine's format at each weight update (Section~\ref{sec:weight-update}).

\begin{itemize}
\item \textbf{MXFP8} shares one UE8M0 scale across 32 consecutive E4M3 values, and \miles{} keeps that format through rollout, the forward pass, and both the weight-gradient and data-gradient GEMMs, holding the configured exceptions in BF16.
\item \textbf{NVFP4} scales activations \emph{per token}, which keeps quantization artifacts from depending on how a batch is composed. \miles{} quantizes the gate and up projections \emph{together}, so the fused rollout GEMM uses one outer weight scale. Tensors the recipe does not cover stay in BF16, and rollout keeps a BF16 KV cache.
\item \textbf{Both recipes} share a bit-exact quantizer, so the training and rollout kernels see identical quantized values, apart from the per-tensor exceptions that stay in BF16.
\end{itemize}

It is worth mentioning that both recipes hold a few tensors in BF16 rather than quantizing them, and those per-tensor exceptions exist because the contraction axis of some tensors, such as the final transformer layers, the shared experts, and the projections in multi-head latent attention, does not line up with a one-dimensional scaling block. \miles{} therefore lets a run override precision by layer range and tensor name at each of the four stages in the contract. An override must be applied at all four aforementioned stages.

In addition to the contract, \miles{} offers two optional refinements to the NVFP4 recipe, both selected through environment variables rather than command-line flags, and the two differ in scope. The first, \emph{dequantized backward}, touches only the training side: it runs the backward GEMMs in BF16 on operands dequantized from the NVFP4 values the forward pass used, trading throughput for gradient stability while leaving the quantized values themselves unchanged. The second, \emph{four-over-six}~\citep{fourover6}, changes how each NVFP4 block is quantized. Quantizing a block maps its largest value onto one of the magnitudes FP4 can represent, and the two candidates are 4 and 6; four-over-six tries both and keeps whichever quantizes that block with less error. Because it changes the quantized values, four-over-six belongs to the contract rather than to the backward pass: the recipe enables it in the Transformer Engine kernels the trainer uses and in the FlashInfer kernels SGLang uses alike, so that checkpoint conversion, the forward pass, weight export, and rollout all quantize a block the same way.

We verify these recipes by comparing the log-probabilities that SGLang and Megatron-LM assign to the same sampled tokens. On the configurations measured so far, the reward curves track the BF16 baseline closely, and the rollout time is reduced significantly. Our evidence covers all tested configurations in \miles{}, but we note that the same quantization might behave differently on different models. MXFP8 and NVFP4 remain Beta, and Table~\ref{tab:precision-formats} shows the models each format has been tested on.

\subsection{Memory Efficiency and Disk Offload}
\label{sec:memory}

Memory capacity limits a training run when a model's weights, gradients, and optimizer state exceed the GPU's available high-bandwidth memory (HBM). If HBM is unable to accommodate everything, the training actor must then move certain states away from the GPU so the training step can fit. When a rollout engine shares the same set of GPUs, offloading is also necessary in order to free capacity for rollout between training steps.

\miles{} offers two offloading mechanisms that act at different times. The first mechanism, evicting the paused actor (Section~\ref{sec:offload-actor}), moves the entire training process off the GPU between training steps: its weights, gradient buffers, and optimizer state leave together, allowing another process (e.g.~a rollout engine) to use the same memory while the actor waits. The second mechanism, streaming the optimizer state (Section~\ref{sec:stream-optimizer}), keeps optimizer state off the GPU during the training step and fetches it back one bucket at a time when the optimizer update needs it. Because the forward and backward passes never read the optimizer state, the memory it would otherwise hold stays free while the actor computes gradients. The two mechanisms compose: a run can offload the actor between steps and stream optimizer state during each step. In addition to these two mechanisms, \miles{} also affords further memory savings that this report does not detail.

\subsubsection{Offloading the Paused Actor}
\label{sec:offload-actor}

Offloading moves a paused training actor's state off the GPU, then restores it when training resumes. In a colocated run, the rollout engine shares the training actor's GPUs (Section~\ref{sec:fully-async}), so \miles{} evicts the actor for exactly the interval in which the engine generates, through \flag{offload-train}.

The Megatron backend transfers the actor's state at the allocator level rather than tensor by tensor, so it never has to know which tensor occupies which bytes. Weights, gradient buffers, and optimizer state leave as one block when the actor pauses and return as one block when it resumes. Pinned host memory is the default and faster destination. When the model is too large for host RAM to hold a full copy, the block goes to node-local disk instead. On the disk path, \miles{} streams the actor to per-rank files through a fixed-size pinned staging buffer, so host memory stays bounded no matter how much is offloaded. The FSDP backend attains the same result by moving the model and optimizer to host memory instead of working through the allocator, but it can offload only to host memory: unlike the Megatron path, it cannot save the block to node-local disk (Table~\ref{tab:backends}).

The default offloading behavior is determined by whether another process needs the training GPUs while the actor is idle. In a colocated placement, the rollout engine needs the training GPUs between the training steps, so \miles{} evicts the training actor by default. Under a disaggregated placement, since the training actor and the rollout engines have separate pools of GPUs, eviction is off by default. PPO (Section~\ref{sec:objective}) is a special case because it requires both an actor and a critic, and \miles{} always colocates actor and critic on the same set of GPUs. Therefore, when the training algorithm is PPO, \miles{} evicts the actor by default whenever a critic is using the GPUs, regardless of placement.

\subsubsection{Streaming the Optimizer State}
\label{sec:stream-optimizer}

The optimizer state usually occupies more HBM than anything else in a large run: the FP32 master weights and two Adam moments together use 12 bytes per parameter. To save memory, \miles{} offers optimizer state streaming through \flag{stream-optimizer-state-to-disk}, which divides model parameters into buckets, and stores one file for all the optimizer state needed to update each bucket of parameters. When the actor updates a bucket of parameters, it loads the corresponding file into HBM, updates the parameters, and then evicts the loaded state buffers from GPU memory. In this way, the GPU doesn't have to hold the full optimizer state at any given time.

Not every run requires optimizer state streaming. For example, data parallelism (DP) already shards the optimizer state across ranks, and optimizer state streaming becomes unnecessary if each sharded optimizer state is small enough to fit into HBM. However, note that the GLM-5.2 reference run in Section~\ref{sec:case-study} still needs optimizer state streaming despite DP = 4, because the sharded optimizer state per DP rank is still too large to be accommodated by HBM.

Streaming optimizer state saves GPU memory at the cost of slower weight update. We can reduce such a cost by lowering the precision of the streamed Adam moments from FP32 to BF16. Note that the master weights have to be saved in FP32 in any case.

When combined with actor offloading, streaming reduces the time it takes to offload an actor, because an optimizer state already on disk does not need to move when the actor is offloaded. On Qwen3-30B-A3B, streaming optimizer state cuts the actor's offloading from 24\,s to 5.2\,s, and its reloading from 8.9\,s to 1.3\,s.

There are, however, a few caveats that are worth mentioning.

\begin{itemize}
\item \textbf{A streamed run resumes only on the same parallel layout.} Each file follows the DP shard of the rank that wrote it, so changing the parallelism triggers a layout assertion.
\item \textbf{A streamed run cannot resume a checkpoint written without streaming.} The streamed files are the only optimizer state \miles{} reads, so \miles{} refuses the resume rather than silently restarting Adam from step zero if the streamed files are not found.
\item \textbf{Checkpoint saves take noticeably longer.} \miles{} copies the streamed state into the checkpoint directory synchronously, outside the asynchronous save path, so the copy may block training.
\end{itemize}

\subsection{Two Training Backends}
\label{sec:backends}

The backend determines how \miles{} partitions and manages the model, while the rest of the training loop uses the same interface. \miles{} provides two implementations: NVIDIA Megatron-LM~\citep{megatron} and PyTorch FSDP~\citep{fsdp}. A single launch flag selects the backend that owns the model on the GPU. Each backend saves checkpoints, runs training steps, reports its parallelism configuration, publishes weights to the rollout engines, and pauses or resumes the training actor (Section~\ref{sec:offload-actor}).

Megatron-LM is the default backend, and is embedded in most shipped recipes, because it offers a more comprehensive set of parallelisms. Those axes make Megatron-LM the practical choice for large MoE models and jobs spanning multiple GPU racks.

FSDP is a sound choice when direct Hugging Face loading matters more than model parallelism. FSDP loads a Hugging Face directory as-is, with no conversion step and no architecture flags to write, making it the shorter path to a first run. That direct path matters most when a researcher brings up a new architecture, checks trainer numerics against the Hugging Face reference, or trains a model that fits under data parallelism alone. A few shipped recipes, including Qwen3 and Nemotron variants, use FSDP. For architectures that need small corrections, the FSDP backend registers them as adaptation specs rather than forking the Hugging Face model code. 

Table~\ref{tab:backends} sets out the remaining differences between the two training backends.

\begin{table}[t]
  \centering
\small
  \begin{tabularx}{\textwidth}{@{}l >{\raggedright\arraybackslash}X l@{}}
    \toprule
     & \textbf{Megatron-LM} & \textbf{FSDP} \\
    \midrule
    Model splitting          & TP $\times$ PP $\times$ CP $\times$ EP $\times$ ETP, plus DP & Replicate $\times$ shard \\
    Model input              & Megatron distributed checkpoint, or Hugging Face via Bridge & Hugging Face directory, as-is \\
    Checkpoints written      & Megatron distributed checkpoint & PyTorch Distributed Checkpoint \\
    Activation recompute     & Megatron recompute settings      & Gradient checkpointing \\
    Optimizer on CPU         & $\checkmark$                     & $\checkmark$ \\
    Offload beyond host RAM  & $\checkmark$ (\S\ref{sec:memory})& $\times$ \\
    Attention backend        & Chosen by Megatron Core          & Selectable \\
    LoRA (\S\ref{sec:lora})  & $\checkmark$                     & $\times$ \\
    \bottomrule
  \end{tabularx}
\caption{The two training backends. Megatron-LM exposes model-parallel axes, while the current FSDP backend uses data-parallel sharding; a $\times$ marks what that backend does not yet implement, not a limit of FSDP itself.}
  \label{tab:backends}
\end{table}

Note that the Megatron-LM backend does not require offline conversion: Megatron-LM can read a Hugging Face directory directly through Megatron Bridge~\citep{megatron_bridge}. The Megatron-checkpoint entry in Table~\ref{tab:backends} is therefore the default input path. A Megatron-LM checkpoint is parallelism-agnostic once written, so a run can change its parallel layout later without reconverting.

\subsection{The Objective and Rollout--Training Corrections}
\label{sec:objective}

\miles{} defines a training run's objective through two layers: an advantage estimator and a typed loss interface. Precision (Section~\ref{sec:low-precision}), memory (Section~\ref{sec:memory}), and the choice of backend (Section~\ref{sec:backends}) usually do not interfere with what the run optimizes, so the run can choose its objective independently of those three properties. Five advantage estimators are shipped with \miles{}: GRPO~\citep{grpo} and GSPO~\citep{gspo}; REINFORCE++~\citep{reinforcepp} in plain and baseline-relative forms; and PPO~\citep{ppo} with a learned value function. The typed loss interface provides one protocol with policy, value, and supervised variants, plus a hook for a user-supplied one. Such an interface allows supervised fine-tuning to reuse the same trainer without an attached rollout engine (Section~\ref{sec:recipes}).

Note that the trainer also has to make corrections for train-rollout mismatch. Since SGLang and Megatron use different kernels, different precisions, and different batching, the two sides disagree even on a trajectory sampled from the current weights, so even a fully synchronous run exhibits train-rollout mismatch. Sections~\ref{sec:tito} and~\ref{sec:r3} remove two structural causes of this mismatch, and Section~\ref{sec:zero-kl} removes the numerical remainder for the configurations it covers. Any remaining difference deviates the importance ratio $r = \exp(\log \pi_{\text{train}} - \log \pi_{\text{rollout}})$ from one, and \miles{} offers two corrections that act on that ratio differently:

\begin{itemize}
\item \textbf{Truncated importance sampling (TIS)} clamps $r$ to a configured interval and uses the clamped value as a per-token weight on that token's policy-gradient loss, so a token with an extreme ratio is damped rather than dropped. The default interval is $[0, 2]$, which acts on the upper tail alone: a ratio above 2 is pulled down, and nothing is pulled up, because a ratio cannot fall below zero.
\item \textbf{Clip-or-pop} sets the per-token weight to zero for any token whose ratio falls outside the same interval, which removes the token's contribution to the gradient exactly as loss-masking would, and passes the unclipped ratio through for tokens inside it. An outlier token is therefore dropped rather than damped.
\end{itemize}

Under either correction, \miles{} reports the same three metrics: the pre-clamp ratio, the clipped fraction, and the mean absolute deviation $|r - 1|$.

\section{Weight Update}
\label{sec:weight-update}

Weight synchronization follows a simple contract: the trainer prepares new weights, transfers them to every rollout engine, and then the rollout engines start to generate under the updated policy. \miles{} performs that handoff on a configurable cadence, after every training step by default. When training and rollout occupy different GPUs, the transfer can become a major pipeline bottleneck, and at frontier scale it can dominate the step: a full NCCL broadcast of Kimi K2 1T-A32B~\citep{kimik2} takes almost a minute. To allow users to select the most efficient way to synchronize model weights, \miles{} provides three weight transports that differ only in how a prepared bucket of weights reaches the rollout engines  (Table~\ref{tab:weight-update}).

\begin{table}[t]
  \centering
\small
  \begin{tabular}{@{}l l l@{}}
    \toprule
    \textbf{Transport} & \textbf{Transfer path} & \textbf{Applicable when} \\
    \midrule
    Broadcast (default)
      & NCCL broadcast to every rollout rank
      & Ranks share an NCCL fabric \\
    Peer-to-peer (\S\ref{sec:p2p})
      & RDMA writes into rollout-rank memory
      & Direct rank-to-rank reachability \\
    Disk-delta (\S\ref{sec:disk-delta})
      & Changed bytes published to shared storage
      & No shared fabric, or transfer dominates \\
    \bottomrule
  \end{tabular}
\caption{The three weight-synchronization transports. All deliver the same converted weights but differ in the connectivity they assume and in how transfer volume scales with the fleet. When training and rollout are colocated on the same GPUs, the handoff is local and no transport is involved.}
  \label{tab:weight-update}
\end{table}

\subsection{The Shared Bucketed Pipeline}
\label{sec:weight-pipeline}

Broadcast and peer-to-peer (P2P), two of the three weight transports in Table~\ref{tab:weight-update}, use one preparation pipeline: the pipeline prepares each bucket once, then hands it to the selected transport. The two transports differ only in how a filled bucket leaves the trainer and which ranks send it. The Megatron backend owns the preparation pipeline and both transports. The FSDP backend instead uses a simpler updater that buckets the model's state dictionary and reaches rollout engines on separate GPUs by broadcast alone.  This section covers the Megatron backend only.

The preparation pipeline batches weights so \miles{} calls the rollout engines once per bucket rather than once for each of a large model's many thousands of tensors. For an ordinary weight, \miles{} all-gathers the Megatron tensor-parallel shards within each pipeline stage, converts the result to the Hugging Face names and layout SGLang expects, and appends it to a fixed-size buffer, 512\,MB by default. Once the buffer fills, \miles{} hands the whole bucket to the selected transport, an operation we call a \emph{flush}. Broadcast then sends the bucket's tensors as one batch of asynchronous NCCL broadcasts, whereas P2P writes them into rollout-rank memory over RDMA.

Routed expert weights require a second pass through the same pipeline, which we call the \emph{expert pass}, because expert parallelism gives each rank a different subset of the experts rather than a shard of one shared tensor. \miles{} therefore fills the expert bucket with shards gathered across tensor parallelism alone and defers the expert-parallel gather and the Hugging Face conversion until the bucket is flushed. Deferring those operations changes bucket sizing: the gather at flush time collects every rank's experts, so the flushed bucket is the accumulated bytes times the expert-parallel degree. To make a flushed expert bucket land at about the intended size, \miles{} multiplies the accumulated bytes by that degree before comparing them against the buffer size. \miles{} runs both passes for every model, whether or not the architecture uses a mixture of experts. A dense model has no routed expert weights, so the expert pass finds nothing to move and issues no collective. When a model's experts sit on every rank, the expert-parallel degree is one, so the multiplier leaves the expert threshold identical to the ordinary one.

The pipeline scales to large models split across many pipeline-parallel stages because \emph{each stage moves its own weights independently}. Under broadcast, each stage has its own NCCL group; under P2P, each stage writes a disjoint parameter set. However, there are two mechanisms that still serialize stages: broadcast takes a single shared lock during each bucket flush, and \miles{} inserts barriers across trainer ranks at phase boundaries (after generation pauses, before/after the expert pass, and after generation resumes). A slow stage therefore stalls all others at those points.

\subsection{Peer-to-Peer Weight Transfer}
\label{sec:p2p}

Peer-to-peer (P2P) transfer\footnote{\url{\docsurl/advanced/p2p-weight-transfer}}~\citep{miles_p2p} treats each weight update as direct writes from training ranks (sources) to rollout ranks (targets). While a broadcast update sends one copy from one sender to every rollout rank, even when some rollout ranks do not need those shards (Table~\ref{tab:weight-update}), P2P transfer lets multiple sources send concurrently, and each source writes only the shards its assigned targets need directly into target memory over remote direct memory access (RDMA). The number of sources that actually send is therefore whichever fleet is smaller: if the sources do not outnumber the targets, every source sends, and if they do, the surplus sources sit idle for that transfer.

Three implementation details make P2P weight transfer practical.

\begin{itemize}
\item A \textbf{transfer plan} assigns each training rank the rollout ranks it will write to. The first $\min(\text{sources}, \text{targets})$ ranks map one to one, and \miles{} spreads the remaining targets evenly over the sources, which limits the number of RDMA sessions any one sender maintains.
\item A \textbf{CPU-resident model replica} mirrors the parallelism layout of the rollout ranks assigned to a sender. \miles{} builds a full SGLang model in CPU memory, allocating nothing on the GPU, purely so that it can call that model's own weight-loading functions. Those functions are the ones a rollout engine would use to place a tensor on a given rank, so running them on the host reshapes each gathered tensor into exactly the layout its target expects, and the sender never has to reimplement SGLang's sharding rules. 
\item A \textbf{single shared pinned buffer} stages every outgoing write. \miles{} registers the buffer for RDMA once and reuses it for every target engine, keeping host memory constant as the number of engines grows. \miles{} groups targets by engine rank rather than by engine, and every group but the last waits for its writes to finish before the next group refills the buffer. The last group's writes go to a background thread pool instead, where they overlap the next bucket's preparation.
\end{itemize}

\begin{table}[t]
  \centering
\small
  \begin{tabular}{@{}l r r r r@{}}
    \toprule
    \textbf{Model} & \textbf{Nodes/side} & \textbf{Broadcast} & \textbf{P2P} & \textbf{Change} \\
    \midrule
    Qwen3-30B-A3B~\citep{qwen3}   & 2  & 2.67\,s  & 2.16\,s & $-19.1\%$ \\
    GLM-5 744B-A40B~\citep{glm5}  & 16 & 58.30\,s & 8.48\,s & $-85.5\%$ \\
    Kimi K2 1T-A32B~\citep{kimik2} & 32 & 53.28\,s & 7.23\,s & $-86.4\%$ \\
    \bottomrule
  \end{tabular}
\caption{Time per weight update, P2P against NCCL broadcast, on H100 clusters with a 1\,GB transfer bucket, averaged over steady-state steps and timed from the end of the generation pause to the return of the update call. Node counts are per side, with trainer and rollout fleets of equal size. The Kimi K2 times include about 884\,ms of on-GPU requantization that its checkpoint requires after every transfer. The advantage grows with fleet width rather than model size, and appears already at two nodes per side.}
  \label{tab:p2p-scaling}
\end{table}

P2P pays off in proportion to how wide the rollout and trainer fleets are. For $M$ source ranks at pipeline depth $p$ and a target expert-parallel degree $e$, P2P draws on roughly $M/p$ times more aggregate sender bandwidth, while each target receives about $e$ times less data. However, we note that \textbf{on single-node deployments P2P is slower than broadcast}, by up to about 70\% in our measurements. This is because a single node gives the senders no extra aggregate bandwidth to draw on, and P2P still pays for host-side re-sharding and pinned-memory staging on every update. As a result, we make broadcast the default transfer, and P2P is only useful when the trainer fleet and the rollout fleet each span more than one node.

Model coverage is limited for P2P weight transfer because a sender can re-shard weights only when the architecture carries a unified weight-name mapping between Megatron and SGLang. Current mappings cover the Qwen2 and Qwen3 dense families, Qwen3-MoE, the GLM4-MoE families, and DeepSeek-V3 and V3.2 derivatives.

\subsection{Disk-Delta Updates}
\label{sec:disk-delta}

Consecutive RL steps change only a small fraction of the model's bytes, so sending just the changed bytes costs far less than sending all of them. The transport that exploits this is disk-delta updates. With disk-delta updates, every rollout host starts from a shared base checkpoint. At each update the trainer writes the changed bytes to a shared filesystem, together with a reference to the base they apply to, and each rollout host reads them and patches its own local copy of the checkpoint. Broadcast and P2P both require the trainer to reach engine memory directly, over NCCL or RDMA (Sections~\ref{sec:weight-pipeline} and~\ref{sec:p2p}); disk-delta needs only a filesystem that both sides can see, which is what makes it usable when no such fabric exists.

Under disk-delta updates, \miles{} still manages or attaches to each SGLang engine, and it writes each new policy to storage as a numbered version. 

\begin{enumerate}
\item On the first update, \miles{} publishes nothing: it captures a host-side snapshot of the base checkpoint, and each rollout host materializes that same base locally.
\item At each subsequent update, trainer ranks gather their tensors under canonical Hugging Face names and compare the bytes against the snapshot.
\item Trainer ranks then publish a new version directory holding the compressed changed bytes, together with an index recording the version, its base version, the delta encoding, and the digest algorithm. \miles{} writes every file atomically and writes the index last, so a rollout host sees a version only once every file is in place.
\item Each rollout host pulls the version and patches its own host-local checkpoint.
\item \miles{} pauses generation, reloads the patched checkpoint into SGLang, advances the engine's weight version, and resumes generation.
\end{enumerate}

Note that only step 5 pauses generation, which is what keeps disk-delta affordable. Steps 1 through 4 only write and read files, and can proceed while the rollout engines are still running in a fully asynchronous run (Section~\ref{sec:fully-async}).
Also note that nothing in the design requires the trainer and the rollout hosts to sit on the same machine, since they need only a filesystem in common.

\miles{} offers two encodings for the weight delta:

\begin{itemize}
\item \textbf{XOR} (the default) stores the bytewise difference. XOR is the more compact of the two and the faster to produce, but it is an \emph{involution}: applying the same delta twice restores the previous weights rather than failing. An XOR delta must therefore be applied exactly once, and only against its declared base.
\item \textbf{Overwrite} stores the changed positions together with their new absolute values. Overwrite writes more bytes than XOR, but it is idempotent: applying it twice gives the same result as applying it once, so a retry or a duplicate does no harm.
\end{itemize}

Both encodings operate on raw bytes and know nothing about the tensors those bytes encode, so the base and the exported policy must agree on tensor names, dtypes, shapes, and byte layout. To ensure correctness, \miles{} checksums each tensor of the \emph{resulting state} rather than the delta. If the base a version records does not match the checkpoint a host actually holds, or if a checksum fails, \miles{} stops the update before any engine reloads, so no engine ever generates from a partly applied set of weights.

Backend support limits disk-delta to Megatron: an FSDP run keeps the broadcast updater of Section~\ref{sec:weight-pipeline} whatever the transport flag says. \miles{} also refuses to start a run that combines disk-delta with colocation, LoRA, or prefill--decode disaggregation. On current \texttt{main}, \miles{} addresses one endpoint for each engine and carries no path to an external rollout service behind a single endpoint, an integration the documentation describes as forthcoming.

\subsection{Pausing Generation, and Checking the Result}
\label{sec:weight-verify}

After the updated weights are synchronized on the rollout engines, two questions still remain: what happens to requests in flight when the weights change, and how does a run confirm that the engines received the trainer's weights?

With the Megatron backend, \miles{} offers three options to handle in-flight rollout requests: a run can abort a request outright, leave it in place while applying the update around it, or retract it so that it rolls back and resumes against the new weights. Retraction is the default. Note that the session server in Section~\ref{sec:tito} cannot tolerate a session's token history being torn down mid-trajectory, so whenever session server is enabled, \miles{} refuses to start a run with the abort mode. With the FSDP backend, in-flight rollout requests are always retracted.

\miles{} ships an opt-in check that confirms the rollout engines hold the weights the trainer sent. The check runs once, at the start of training, and compares each engine's weights against the trainer's own. \miles{} first fills the engine's tensors with random values and only then runs the first weight update, so any tensor the transport failed to write still holds that noise when the check reaches it; a silent omission therefore cannot pass unnoticed. A run can point the check at the target model, at the draft model that multi-token prediction adds, or at both. Tensors that legitimately live only on the rollout side, such as inference-only cache scales, are skipped, and a run can name further tensors to skip. The check demands bit-exact equality by default, but a run may instead admit the rounding error of a quantized round trip, and \miles{} derives that tolerance from the quantized format rather than from a number the user supplies.

Note that the tensor-by-tensor check should only take place in debugging and validation runs, so \miles{} enables it automatically only in its own continuous-integration tests. We have used the check to validate peer-to-peer transfer (Section~\ref{sec:p2p}) across model families, on GPU fleets from one node up to eight.

\section{Other Post-Training Recipes}
\label{sec:recipes}

The RL loop of Sections~\ref{sec:rollout} through~\ref{sec:weight-update} updates every parameter of one model, using a reward computed on the trajectories that the model itself generated. Not every post-training job takes that shape. \miles{} therefore treats the rollout engines, the trainer, and the weight-update paths as shared components rather than RL-specific machinery, and composes three further recipes from them, each of which changes one part of the loop and leaves the rest intact. LoRA RL (Section~\ref{sec:lora}) changes which parameters the trainer updates. On-policy distillation (Section~\ref{sec:opd}) changes where the training signal comes from, by letting a teacher model's judgment supplement or replace the reward. True-on-policy alignment (Section~\ref{sec:zero-kl}) changes the numerics rather than the algorithm, so that the rollout engine and the trainer assign exactly the same probability to every sampled token. Supervised fine-tuning also runs on the same trainer, with a supervised loss and no rollout engine (Section~\ref{sec:objective}). Since it involves little that is specific to \miles{}, this report does not discuss it further. This section examines the three recipes in that order.

\subsection{LoRA RL}
\label{sec:lora}

Full-parameter RL pays for the whole model at every step: the trainer holds gradients and optimizer state for every parameter, and each weight update moves every parameter to the rollout engines. Low-rank adaptation, or LoRA~\citep{lora}, avoids most of that cost by keeping the base model frozen and learning a small \emph{adapter} instead: a pair of low-rank matrices attached to each selected module, whose product acts as an additive correction to that module's frozen weight. The adapter's size depends on its rank and on which modules it wraps, and it is typically a small fraction of the base model. In \miles{}, the adapter becomes the loop's unit of work.\footnote{\url{\docsurl/advanced/lora}} The trainer updates it, the weight-update path synchronizes it, and SGLang applies the newest copy during rollout while the base checkpoint stays resident on every engine.

Because only the adapter changes, LoRA lightens two of the three stages of the loop, the training step and the weight update.

\begin{itemize}
\item \textbf{Less time and memory per training step.} The trainer computes gradients and optimizer updates only for the adapter matrices. The optimizer state, which Section~\ref{sec:stream-optimizer} identified as the largest consumer of GPU memory in a full-parameter run, shrinks to the adapter's parameters, and each step does less arithmetic, moves less memory, and communicates less across ranks.
\item \textbf{Less time per weight update.} \miles{} synchronizes the updated adapter rather than the whole model, so each update moves a small fraction of the model weights. A colocated job, in which the trainer and the engine share a GPU, passes the adapter through inter-process communication (IPC): the trainer serializes the adapter tensors and hands the engine process a handle to them, so nothing crosses the network. A disaggregated job instead broadcasts serving-ready adapter tensors to the remote SGLang engines over NCCL (Section~\ref{sec:weight-pipeline}). The P2P and disk-delta transports (Sections~\ref{sec:p2p} and~\ref{sec:disk-delta}) do not carry adapters.
\end{itemize}

Whether a model supports LoRA is a contract among three components. The Megatron adapter implementation must build and wrap the training model, \miles{} must map and export the adapter's module names correctly, and SGLang must allocate and apply the same adapter modules at serving time. A checkpoint whose modules carry the same names as a supported model's, such as the usual attention and MLP projections, can still fail one of the three on its expert, linear-attention, or model-specific projections, so a name match alone does not establish support. \miles{} therefore ships validated recipes rather than an allowlist, covering dense and MoE models including Qwen2.5~\citep{qwen25}, Qwen3~\citep{qwen3}, gpt-oss~\citep{gptoss}, Kimi K2.5~\citep{kimik25}, GLM-5/5.1/5.2~\citep{glm5,glm51_blog,glm52_blog}, Qwen3.5/3.6~\citep{qwen35,qwen36}, and Inkling~\citep{inkling}. Currently, LoRA training runs on the Megatron backend only (Table~\ref{tab:backends}). LoRA also composes with the ``BF16 train, FP8 serve'' arrangement of Section~\ref{sec:low-precision}: the validated GLM-5.2 recipe trains the adapter in BF16 while SGLang serves an FP8 copy of the base, and the adapter tensors themselves stay unquantized.

\miles{} can also train several adapters at once against one shared base model, through an experimental multi-LoRA path. The rollout engines keep a single copy of the base and serve every adapter from it, and the trainer likewise holds every adapter beside one base model, so a training batch can mix samples from several adapters. Each registered adapter carries its own dataset, reward, optimizer state, and step count, and after each step \miles{} sends the engines only the adapters whose optimizer stepped. Multi-LoRA supports disaggregated rollout only. The operation-driven backend and the Tinker-compatible frontend described in the LoRA documentation remain open pull requests rather than released features.

Miles-Diffusion (Section~\ref{sec:diffusion}) trains LoRA adapters as well, although its rollout engines merge the adapter into the base weights on arrival rather than serving it separately.

\subsection{On-Policy Distillation (OPD)}
\label{sec:opd}

\begin{figure}[t]
  \centering
  \includegraphics[width=0.85\textwidth]{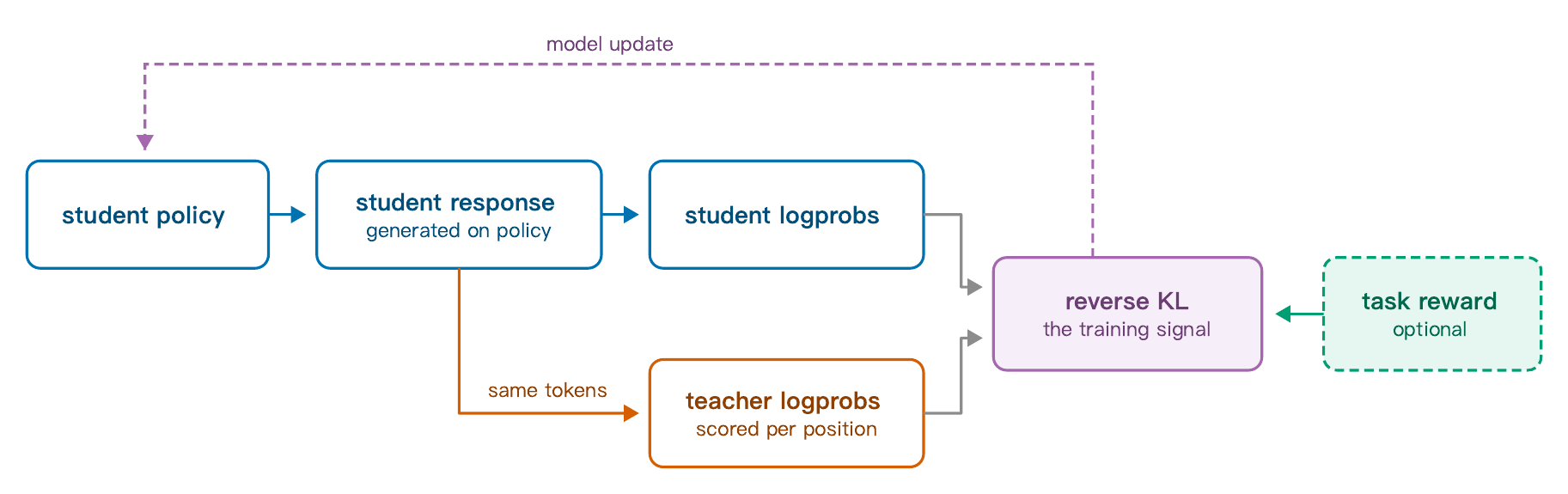}
\caption{On-policy distillation. The student generates a response, the teacher scores the same tokens position by position, and the difference between the two log-probabilities, a per-token reverse-KL estimate, drives the update, optionally alongside a task reward.}
  \label{fig:opd}
\end{figure}

On-policy distillation (OPD)~\citep{tml_opd} trains a student model on its own trajectories while a stronger teacher model supplies the learning signal.\footnote{\url{\docsurl/advanced/on-policy-distillation}} Conventional distillation trains the student on text the teacher wrote, which the student would rarely produce on its own. OPD instead lets the student generate, and then asks the teacher how likely it would have been to choose each token the student actually chose. The per-token signal is the difference between the two log-probabilities of the sampled token $x_t$, namely $\log \pi_{\text{student}}(x_t) - \log \pi_{\text{teacher}}(x_t)$, which is a one-sample estimate of the reverse Kullback--Leibler divergence at that position. The signal is positive where the student favored the token more strongly than the teacher, and negative where the teacher favored it more. Figure~\ref{fig:opd} shows the scheme.

\miles{} folds that signal into the advantage rather than the loss. After the configured advantage estimator (Section~\ref{sec:objective}) has computed token-level advantages, \miles{} subtracts the divergence estimate, scaled by a coefficient, from each token's advantage, so a token the student favored more strongly than the teacher is penalized and a token the teacher favored more strongly is encouraged. The policy-gradient update then proceeds unchanged. Distillation therefore composes with GRPO~\citep{grpo}, PPO~\citep{ppo}, and the other estimators rather than replacing them, and a run may keep the task reward alongside the teacher's signal or set the reward to zero and distill alone. Both log-probabilities in the estimate are fixed inputs, recorded at rollout time or produced by a separate teacher pass, so the penalty acts as a dense per-token reward rather than as an extra loss term.

The sampled token alone gives a one-sample estimate of the divergence at each position. \miles{} therefore also offers a top-$K$ variant, following \citet{rethinking_opd}, that sits between the one-sample estimate and the divergence over the full vocabulary: the teacher scores a set of candidate tokens at each position, such as the student's $K$ most likely tokens or the intersection of the student's and the teacher's top-$K$ sets, and the estimate becomes a weighted sum of the differences over that set.

Where the teacher runs determines when \miles{} obtains its log-probabilities. With a \emph{served} teacher, an external SGLang server scores each finished trajectory during rollout, and the teacher's log-probabilities travel to the trainer with the trajectory. The teacher may then have a different architecture from the student, or be too large to load beside it, but it must share the student's tokenizer, because scoring is performed on the student's token IDs. With an \emph{in-process} teacher, Megatron loads a second model of the same architecture next to the student and scores each batch in a dedicated forward pass during the training step. The top-$K$ variant is available with a served teacher only. A run may also register several served teachers and route each prompt to one of them by a tag in the prompt's metadata, so that, for example, a mathematics specialist scores mathematics prompts and a coding specialist scores coding prompts.

Distillation alone can shorten a student's responses by more than half without a reliable change in accuracy. In the documented Qwen3.5-35B-A3B run~\citep{miles_opd}, the teacher is a copy of the same model improved by five steps of RL with a verifiable reward, the student starts from the base checkpoint, and the task reward is set to zero, so the reverse-KL penalty supplies the entire training signal. Over five steps, the response length on held-out DAPO~\citep{dapo} prompts falls from 14{,}070 to 6{,}132 tokens, while accuracy moves from 84.0\% to 85.2\%. That 1.2-point movement lies inside the evaluation's standard error of roughly 1.6 points. The supported conclusion is therefore a 56\% reduction in response length with no reliable change in accuracy, not a demonstrated improvement on the benchmark.

\subsection{True-On-Policy Alignment}
\label{sec:zero-kl}

Even when the rollout engine and the trainer hold identical weights, they do not assign identical probabilities to the tokens they score. SGLang and the training backend compute with different attention kernels, different matrix-multiplication kernels, different operator fusions, and different batch shapes, and each difference perturbs the result in the last bits of floating-point arithmetic. Over a long sequence, those perturbations accumulate into a measurable gap between the probability under which a token was sampled and the probability the trainer computes for it, and Section~\ref{sec:objective} describes the corrections that the objective applies to that gap. For the configurations it covers, \miles{} offers a stricter alternative: make the two engines agree exactly, a mode we call \emph{true-on-policy alignment}.\footnote{\url{\docsurl/examples/infra-features/true-on-policy}}

Exact agreement requires every operation on the two sides to produce bitwise-identical results, so \miles{} pins each source of disagreement in turn. Both engines run the same attention kernel, FlashAttention-3~\citep{flashattn3}, whose prefill and decode paths agree bitwise. Both engines also run matrix-multiplication kernels whose output does not depend on how many requests share a batch, a property called batch invariance~\citep{tml_nondeterminism}, because \miles{} cannot make the rollout engine's batches match the trainer's. On the Megatron side, \miles{} swaps Transformer Engine's fused implementation for Megatron's local one, disables the fused rotary-embedding and bias-SwiGLU kernels, and follows a per-model \emph{kernel contract} that pins the remaining operators to the implementations SGLang uses; on the FSDP side, it selects the matching attention implementation. SGLang runs in its deterministic-inference mode, the trainer runs with cuBLAS, Transformer Engine, and NCCL configured for deterministic execution, and when tensor parallelism is in use, the row-parallel linear layers and the all-reduce that follows them are made invariant to the degree of parallelism as well. Finally, the rollout engine re-scores each finished sequence with a prefill pass rather than reporting the log-probabilities its decode kernels produced, so that both sides compute the score with the same shape of computation. In supported configurations, the two engines then produce identical log-probabilities for every sampled token, and the absolute difference that \miles{} reports between them is exactly zero.

Exactness costs throughput, because deterministic and batch-invariant kernels give up some of the optimizations their defaults. In the documented Qwen3-4B-Base run on DAPO~\citep{dapo} prompts, the reward curve under true-on-policy alignment matches the baseline, while rollout takes longer.

The zero-difference result has two limits, model coverage and comparison scope. Current \texttt{main} registers a profile only for the dense Qwen3~\citep{qwen3} 0.6B and 4B variants, under Megatron or FSDP with data, tensor, pipeline, or context parallelism, and \miles{} refuses to start a true-on-policy run for a model outside that profile rather than running it with a partial guarantee. The guarantee itself covers one metric, the log-probability of each sampled token; it does not claim that the two engines agree on every token of the output distribution. Nor does the alignment address the other source of off-policy data, a trajectory generated under weights older than the trainer's, which the data buffer of Section~\ref{sec:data-buffer} bounds separately.

\section{Miles-Diffusion}
\label{sec:diffusion}

Diffusion models are post-trained with RL for the same reason language models are, to align what they generate with a reward, and the loop has the same three stages: an engine generates, a trainer updates, and the new weights return to the engine. Miles-Diffusion\footnote{\url{https://github.com/radixark/miles_diffusion}} carries the design of Sections~\ref{sec:rollout} through~\ref{sec:weight-update} over to image and video diffusion models. In this setting a trajectory is the sequence of denoising steps that turns noise into one image or video. The sglang-diffusion engine generates each trajectory and returns it whole, with the intermediate state and a log-probability recorded at every step, and an FSDP2~\citep{fsdp} trainer re-scores a chosen subset of those steps and optimizes the RL objective on them. The trainer shards the model with FSDP2 and can split a long video sequence across ranks with sequence parallelism~\citep{usp}, a path that the Wan2.2~\citep{wan} recipe exercises. Because the loss, the training-batch preparation, the rollout function, and the reward are all replaceable components, one trainer runs Flow-GRPO~\citep{flowgrpo}, DiffusionNFT~\citep{diffusionnft}, and supervised fine-tuning. Flow-GRPO has a shipped recipe for every supported model, whereas DiffusionNFT and supervised fine-tuning have shipped recipes on SD3.5~\citep{sd3} and Wan2.2 respectively and exist as code paths without a recipe on the others.

Generation, decoding, and reward scoring overlap rather than running as three phases. The engine divides each batch of requests into \emph{microgroups}, each a batch of samples that shares one set of encoder results, and Miles-Diffusion deserializes and scores each microgroup as soon as it arrives while the engine works on the next. Deserialization matters more here than for text, because the trajectory of one video microgroup can be gigabytes of tensors. Miles-Diffusion therefore receives those tensors as raw bytes rather than as base64 text, and unpacks them in a pool of worker processes rather than on the main event loop. On the LTX-2.3 recipe, that path cut rollout time from 157.4 to 87.6 seconds per step and total step time from 321.9 to 252.1 seconds.\footnote{\url{\docsurl/diffusion/advanced/streaming-reward}}

Diffusion RL is unusually sensitive to precision, since the importance ratio compares a log-probability the engine produced with one the trainer produced, and any rounding difference between the two forward passes becomes an update signal made of nothing but noise. Miles-Diffusion therefore adds two controls. A deterministic mode makes the training actor's forward and backward passes repeatable across runs on the same hardware, for the recipes and attention backends that support it: it patches the FlashAttention kernels to run deterministically and rejects attention backends that offer no such switch. Repeatability is what lets the end-to-end tests compare every registered metric of a recipe bit for bit against a committed standard, rather than within a tolerance wide enough to hide a regression. Per-parameter dtype control, a patch to FSDP2's mixed-precision policy, keeps a few precision-sensitive parameters, such as timestep embedders and modulation tables, in FP32 while the rest of the forward pass runs in BF16, and so narrows the training--inference gap itself, the same gap that Section~\ref{sec:low-precision} addresses for language models.

Current \texttt{main} ships recipes for SD3.5~\citep{sd3}, Qwen-Image~\citep{qwenimage}, Wan2.2~\citep{wan}, LTX-2.3~\citep{ltx2}, and Cosmos 3~\citep{cosmos3}, while the MiniMax H3~\citep{minimaxh3} recipe exists only in an open pull request. Every recipe carries an explicit evidence level, and the level applies to the exact script and topology named in the model guide rather than to the model family.\footnote{\url{\docsurl/diffusion/user-guide/recipe-verification}}

\begin{itemize}
\item \textbf{Fully gated.} A complete training curve has been run, and a deterministic end-to-end test runs the recipe itself at least nightly, with every registered metric matching a committed standard exactly.
\item \textbf{Proxy gated.} The same, except that the nightly test runs a documented, scaled-down single-node proxy of the recipe.
\item \textbf{Verified.} A complete training curve has been run, but no deterministic test meets either standard above.
\item \textbf{Not verified.} No complete training curve has been run; smoke tests and short debugging runs do not count.
\end{itemize}

On current \texttt{main}, the SD3.5 and LTX-2.3 recipes are fully gated, the multi-node Wan2.2 recipe is proxy gated, the Qwen-Image and Cosmos 3 recipes are verified, and the two Wan2.2 LoRA recipes are not verified. Stating the level keeps a recipe's presence in the source tree from being read as a claim about how well it works.

\section{Verified Coverage: Models and Hardware}
\label{sec:coverage}

A post-training system is only as useful as the range of models and hardware on which it has actually been run, and a list of names says little unless it also says how each entry was verified. This section reports that range in two parts, model architectures in Section~\ref{sec:day0} and GPU hardware in Section~\ref{sec:hardware}, and each part states where its coverage ends.

\subsection{Day-0 Model Support}
\label{sec:day0}

Supporting a model on the day its weights become public is the strictest test of a post-training system's model coverage, because nothing about the new architecture can be prepared for in advance. \miles{} and SGLang met that test for six frontier models: Kimi K3~\citep{day0_kimi_k3}, DeepSeek-V4~\citep{day0_deepseek_v4}, GLM-5.2~\citep{glm52_blog}, Qwen3.8~\citep{day0_qwen38}, Inkling~\citep{day0_inkling}, and NVIDIA Nemotron 3 Ultra~\citep{day0_nemotron3}. The two projects worked in parallel: SGLang built the inference path for each new architecture, while \miles{} built the RL recipe on top of it. For two of the six, Kimi K3 and the 2.4T-parameter Qwen3.8 mixture of experts, the release-day recipe trained LoRA adapters (Section~\ref{sec:lora}) rather than every parameter, since a frozen base with adapters was what let a model of that size fit the cluster at all. Note that those two recipes still live in open pull requests rather than on current \texttt{main}.

Beyond the release-day set, \miles{} documents recipes for nine model families, spanning dense and MoE models, hybrid attention, and multimodal inputs.\footnote{\url{\docsurl/models}} DeepSeek-V3.2~\citep{deepseek_v32}, Kimi K2.6~\citep{kimik26}, Gemma 4~\citep{gemma4}, and gpt-oss~\citep{gptoss} sit alongside the release-day names, as do earlier Qwen, GLM, and Nemotron generations. However, ``supported'' does not mean one thing across that list. For some checkpoints it means a validated run at full scale; for others, a maintained launcher plus an end-to-end CI test on a reduced-layer slice of the model; for others still, unit tests alone. This report therefore does not treat the list as one uniform claim, and the model page for each checkpoint, rather than this report, records the level that applies to it.

\subsection{Multi-Hardware Support}
\label{sec:hardware}

\miles{} runs the same training loop on NVIDIA and AMD GPUs, though the depth of coverage varies by device.\footnote{The installation guide records the current support status of every NVIDIA and AMD GPU: \url{\docsurl/getting-started/installation}.} On NVIDIA hardware, support spans A100 through GB300. Hopper (H100 and H200) and Blackwell (B200, B300, GB200, and GB300) are production hardware, and Hopper carries most GPU tests in continuous integration. A100 runs the loop with every FP8 feature disabled, lacking FP8 arithmetic. Blackwell features prominently here: the reference run of Section~\ref{sec:case-study} uses 64 GB300 GPUs, and the MXFP8 and NVFP4 recipes of Section~\ref{sec:low-precision} run only on Blackwell.

AMD support is native to ROCm rather than provided through a translation layer~\citep{miles_rocm,miles_rocm_dsv4}, and it covers four Instinct GPUs: MI300X, MI325, MI350X, and MI355X. These GPUs run the same SGLang engine for rollout as the NVIDIA GPUs, but from a separate set of container images, and the shipped AMD launchers cover Qwen3-4B, Qwen3-30B-A3B, GLM-5.2, DeepSeek-V4, and Inkling. Continuous integration runs end-to-end training tests on MI350 runners for every pull request, alongside the NVIDIA runners, and a nightly suite exercises larger MI350 configurations.

Hardware also constrains precision, so a valid run must match its GPU to a supported number format. Section~\ref{sec:low-precision} gives the mapping: MXFP8 and NVFP4 require Blackwell; FP8 blockwise runs on Hopper, Blackwell, and AMD MI350X and MI355X; and A100 uses the BF16 path only. Model support adds a third constraint (Section~\ref{sec:day0}), so a workable run needs a compatible checkpoint, GPU, and number format at once.

\section{Code Quality Principle}
\label{sec:code-quality}

A post-training system accumulates mechanisms quickly, and each one described in this report, from the rollout scheduler (Section~\ref{sec:rollout}) to the weight-update transports (Section~\ref{sec:weight-update}) and the diffusion port (Section~\ref{sec:diffusion}), is a place where the code could have become hard to follow. A single principle organizes \miles{} against that drift: the system should be easy to read and easy to extend.

Readability starts at the training driver, which reads like pseudocode by design. One iteration of the synchronous loop is a handful of calls in sequence: generate a batch of trajectory groups, train on them, save a checkpoint when one is due, release or offload the trainer's GPU memory if the placement calls for it, update the rollout engines' weights, and evaluate when an evaluation is due. Everything that makes those calls work lives behind them, in modules the driver never has to know about.

Extensibility comes from small, typed interfaces at the points where users most often need their own logic: the rollout function, the data source, the reward, the loss, and the correction applied to the importance ratio. Each is selected by naming an import path in a flag, so a user's code is loaded only when a run asks for it, and nothing inside \miles{} has to change. The rollout stack in particular is divided into agent, generation, and rollout layers (Section~\ref{sec:environments}), so an environment or agent framework can replace exactly the layer it needs and reuse the rest. Together, the interfaces and the layer boundaries let a user build a custom RL training loop without forking or rewriting \miles{}.

The principle is enforced rather than merely stated. Formatting and import order are pre-commit hooks that continuous integration reruns on every pull request, and three further hooks ban \miles{}-specific patterns outright, each pointing at the API to use instead; reading Megatron's parallel state directly is rejected, for example, because the two training backends share one parallel-state object. Continuous integration also keeps every test's training metrics across runs and gates each new number against that history, so a slow drift in reward or divergence that no single run would reveal is caught.

\section{Case Study: GLM-5.2 Agentic Training with 64 NVIDIA GB300 GPUs}
\label{sec:case-study}

The preceding sections describe the mechanisms of \miles{} one at a time. This section examines one end-to-end run that exercises most of them at once: \miles{} trains GLM-5.2 744B-A40B~\citep{glm5,glm52_blog} on terminal-use coding tasks with fully asynchronous RL (Section~\ref{sec:fully-async}) across 64 NVIDIA GB300 GPUs, dividing the fleet evenly between generation and training, with 32 GPUs in each role. Section~\ref{sec:case-config} sets out the configuration and the constraints that forced it, and Section~\ref{sec:case-results} reports what the run measured.

\subsection{Configuration}
\label{sec:case-config}

The run must generate with and train a 744B-parameter model at the same time on 64 GPUs, and that constraint decides most of the configuration. The 64 GPUs occupy 16 NVIDIA GB300 nodes of four GPUs each: eight nodes generate trajectories and eight nodes train the model. Each generation node runs one four-GPU rollout engine, which serves an FP8 copy of the model with an FP8 key-value cache while the trainer keeps its weights in BF16, the ``BF16 train, FP8 serve'' arrangement of Section~\ref{sec:low-precision}. A small draft model attached to each engine proposes more than one token per forward pass, a technique called multi-token prediction (MTP).

Memory, rather than throughput, determines how the 32 training GPUs partition the model. Tensor, pipeline, context, and expert parallelism together use all 32 GPUs, leaving a single data-parallel replica, so each rank holds its share of the optimizer state unpartitioned. That share, about 279\,GB per rank, exceeds the GPU's memory on its own, so the run streams it to node-local disk (Section~\ref{sec:stream-optimizer}) as a necessity rather than an optimization. Tensor parallelism stays at degree 2 because, at degree 1, a rank's non-expert weights alone overflow its GPU while the checkpoint loads. Context parallelism then splits each training sequence across four ranks, so each rank holds roughly one quarter of the activations, and pipeline parallelism divides the model's 78 layers over four stages, 18 on the first and 20 on each of the others, because the model's sparse attention shares indices across layers and every stage must start on a layer that computes its own.

Each trajectory in the run is one agent solving one terminal-bench-2 task at a command line inside a sandbox built from the task's official image, and the sandbox is deleted when the episode ends. \miles{} reaches the sandboxes through the OpenEnv connector of Section~\ref{sec:environments}. An episode ends after 30 turns or one hour of wall-clock time, whichever comes first. Each reply may contain at most 8{,}192 tokens, and the maximum sequence length in Table~\ref{tab:case-study} budgets the agent's whole multi-turn session rather than one reply. The task's own test script grades the finished session, and GRPO compares each trajectory with the other seven attempts at the same task to compute its advantage (Section~\ref{sec:objective}), with truncated importance sampling correcting the remaining train--rollout mismatch. Up to 128 trajectories are in flight at once, decoupled from the training batch of 64, and every ten steps the run pauses generation to evaluate on a disjoint held-out set of tasks on the same engines.

\begin{table}[!htbp]
  \centering
\small
  \begin{tabular}{@{}l l@{}}
    \toprule
    \textbf{Setting} & \textbf{Value} \\
    \midrule
    Model                    & GLM-5.2, 744B total / A40B active parameters \\
    Task distribution        & Terminal-bench-2 terminal-use coding tasks \\
    Environment              & OpenEnv with one Daytona sandbox per episode \\
    Hardware                 & 64 NVIDIA GB300 GPUs (32 rollout / 32 training) \\
    Training parallelism     & TP\,2 / PP\,4 / CP\,4 / EP\,8 \\
    Inference parallelism    & Eight DP-attention engines (DP\,4), MTP enabled \\
    Precision                & BF16 training; FP8 weights and KV cache in rollout \\
    Max sequence length      & 65{,}536 tokens per session \\
    Training batch size      & 64 trajectories (8 tasks $\times$ 8 attempts) \\
    Schedule                 & Fully asynchronous (Section~\ref{sec:fully-async}) \\
    \bottomrule
  \end{tabular}
\caption{Configuration of the GLM-5.2 agentic RL reference run.}
  \label{tab:case-study}
\end{table}

\subsection{Results}
\label{sec:case-results}

\begin{figure}[t]
  \centering
  \includegraphics[width=0.95\textwidth]{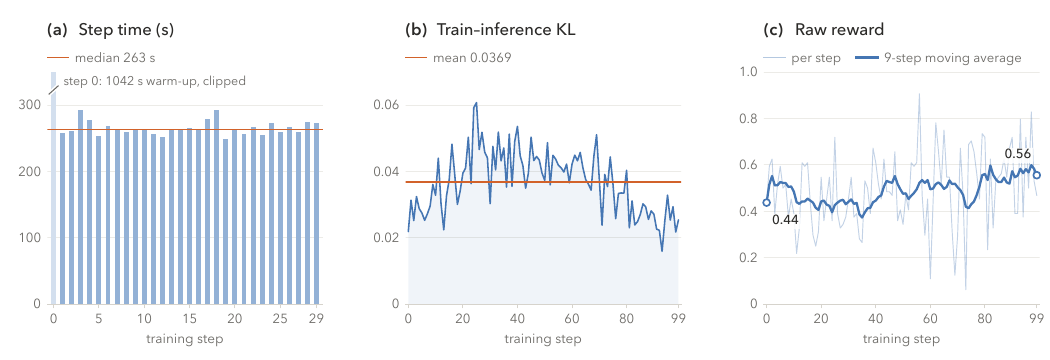}
\caption{Metrics from the GLM-5.2 agentic RL reference run. (a) wall-clock seconds per training step for the first 30 measured steps; the step-0 warm-up value is clipped, and the median is marked. (b) the divergence between the rollout engine's and the trainer's log-probabilities at the sampled tokens, with the mean marked. (c) raw task reward, with a faint line for each step and a bold line for its nine-step moving average.}
  \label{fig:case-study}
\end{figure}

The results come from a single 100-step run under the configuration of Table~\ref{tab:case-study}, and the launch script that reproduces it is in the \miles{} repository.\footnote{\url{\repourl/tree/main/examples/experimental/openenv/glm52_tbench2}} Figure~\ref{fig:case-study} shows three metrics from that run, and three observations about the system follow from it.

\begin{itemize}
\item \textbf{A 744B-parameter model trains on 32 GPUs.} The parallel layout of Table~\ref{tab:case-study} together with the memory mechanisms of Section~\ref{sec:memory} fits the trainer on half of the 64-GPU fleet and leaves the other half to generation.
\item \textbf{The median training step takes 263 seconds.} The median covers the first 30 measured steps in panel (a) of Figure~\ref{fig:case-study}; the 1{,}042-second warm-up value of step 0 is clipped from the plot.
\item \textbf{Generation and training overlap between weight updates.} Sample granularity (Section~\ref{sec:submission}) refills a generation slot as soon as one trajectory finishes, so roughly 90 to 100 requests generate at once across the fleet. The count stays below the limit of 128 trajectories in flight because a trajectory waiting on a tool call holds its slot without generating. Affinity (Section~\ref{sec:fast-rollout}) returns each later turn to the data-parallel rank that already holds its prefix, and the prefix-cache hit rate stays at 96\%.
\end{itemize}

The other two panels assess the run's numerical health and its reward rather than its speed. Panel (b) plots the divergence between the log-probabilities that the rollout engine reported for the sampled tokens and those that the trainer computes for the same tokens. That divergence averages 0.0369 over the 100 steps and ends near its starting value, and the truncated importance sampling of Section~\ref{sec:objective} corrects for it in the update. Panel (c) plots the raw task reward, the score that each task's own test script returns before GRPO centers it within its group; its nine-step moving average rises from 0.438 to 0.556 over the same 100 steps. However, a single run on a single task distribution cannot separate that rise from run-to-run variation, so we report it as an observation rather than as a measured improvement.

\section{Conclusion}
\label{sec:conclusion}

\miles{} brings numerical fidelity, execution efficiency, and extensibility together in one full-stack design for frontier-scale post-training. The rollout stack preserves token-exact multi-turn trajectories and consistent expert routing, while the training stack covers low-precision execution, memory-efficient state management, a choice of two training backends, and corrections for rollout-training mismatch. Weight-update transports then move prepared weights across deployment topologies while preserving one synchronization contract. The same components therefore serve synchronous and asynchronous schedules and several post-training objectives without tying any objective to one deployment.

The architecture also extends beyond the core full-parameter RL loop to LoRA RL, on-policy distillation, supervised fine-tuning, true-on-policy alignment, and diffusion models. Tokenization checks, evidence levels for recipes, and verified model and hardware coverage make the supported scope explicit; current limits include incomplete vision-language session support, early-stage precision formats, and weight-transfer paths that cover only some model families. \miles{} is open source~\citep{miles_repo}, and its major components sit behind small typed interfaces, so a researcher can replace one layer without forking the system.

\section{Contributions and Acknowledgments}
\label{sec:contributions}

The core contributors of \miles{} are Tom Chen, Mao Cheng, Shi Dong, Kangrui Du, Yanbin Jiang, Jiajun Li, Yiming Li, Tao Lin, Yusheng Su, Andy Ye, Yueming Yuan, and Zhichen Zeng. 

We would like to extend our sincere gratitude to Haoguang Cai, Richard Chen, Jiadong Guo, Mathew Han, Lingyan Hao, Kaixi Hou, Nan Jiang, Xinyu Jiang, Zhiyao Jiang, Ziang Li, Mingyi Lu, Qijia Shen, Ying Sheng, Katherine Wang, Yutong Wang, Zhihao Wang, Douglas Yang, Shang Yang, Xinyu Zhang, Chenyang Zhao, Yuzhen Zhou, Banghua Zhu, and Zilin Zhu, as well as to all community contributors of \miles{}\footnote{Full contributors: \url{https://github.com/radixark/miles/graphs/contributors?all=1}} and SGLang\footnote{Full contributors: \url{https://github.com/sgl-project/sglang/graphs/contributors?all=1}} ecosystem, for their invaluable contributions and support.
\vspace{\baselineskip}
\noindent

\bibliographystyle{unsrtnat}
\bibliography{references}

\begin{thebibliography}{65}
\providecommand{\natexlab}[1]{#1}
\providecommand{\url}[1]{\texttt{#1}}
\expandafter\ifx\csname urlstyle\endcsname\relax
  \providecommand{\doi}[1]{doi: #1}\else
  \providecommand{\doi}{doi: \begingroup \urlstyle{rm}\Url}\fi

\bibitem[{THUDM}(2025)]{slime}
{THUDM}.
\newblock {slime: An LLM Post-Training Framework for RL Scaling}.
\newblock GitHub repository, 2025.
\newblock URL \url{https://github.com/THUDM/slime}.

\bibitem[Zheng et~al.(2024)Zheng, Yin, Xie, Sun, et~al.]{sglang}
Lianmin Zheng, Liangsheng Yin, Zhiqiang Xie, Chuyue Sun, et~al.
\newblock {SGLang: Efficient Execution of Structured Language Model Programs}.
\newblock In \emph{Advances in Neural Information Processing Systems
  (NeurIPS)}, 2024.
\newblock arXiv:2312.07104.

\bibitem[Shoeybi et~al.(2019)Shoeybi, Patwary, Puri, LeGresley, Casper, and
  Catanzaro]{megatron}
Mohammad Shoeybi, Mostofa Patwary, Raul Puri, Patrick LeGresley, Jared Casper,
  and Bryan Catanzaro.
\newblock {Megatron-LM: Training Multi-Billion Parameter Language Models Using
  Model Parallelism}.
\newblock \emph{arXiv preprint arXiv:1909.08053}, 2019.

\bibitem[Zhao et~al.(2023)Zhao, Gu, Varma, Luo, et~al.]{fsdp}
Yanli Zhao, Andrew Gu, Rohan Varma, Liang Luo, et~al.
\newblock {PyTorch FSDP: Experiences on Scaling Fully Sharded Data Parallel}.
\newblock \emph{arXiv preprint arXiv:2304.11277}, 2023.

\bibitem[Hu et~al.(2022)Hu, Shen, Wallis, Allen-Zhu, Li, Wang, Wang, and
  Chen]{lora}
Edward~J. Hu, Yelong Shen, Phillip Wallis, Zeyuan Allen-Zhu, Yuanzhi Li, Shean
  Wang, Lu~Wang, and Weizhu Chen.
\newblock {LoRA: Low-Rank Adaptation of Large Language Models}.
\newblock In \emph{International Conference on Learning Representations
  (ICLR)}, 2022.
\newblock arXiv:2106.09685.

\bibitem[Shao et~al.(2024)Shao, Wang, Zhu, Xu, Song, et~al.]{grpo}
Zhihong Shao, Peiyi Wang, Qihao Zhu, Runxin Xu, Junxiao Song, et~al.
\newblock {DeepSeekMath: Pushing the Limits of Mathematical Reasoning in Open
  Language Models}.
\newblock \emph{arXiv preprint arXiv:2402.03300}, 2024.
\newblock Introduces Group Relative Policy Optimization (GRPO).

\bibitem[{Harbor Framework}(2026)]{harbor}
{Harbor Framework}.
\newblock {Harbor: An Agent Evaluation and Training Harness}.
\newblock GitHub repository, 2026.
\newblock URL \url{https://github.com/harbor-framework/harbor}.

\bibitem[{NVIDIA}(2026{\natexlab{a}})]{nemogym}
{NVIDIA}.
\newblock {NVIDIA NeMo Gym: Environments for Agentic Post-Training}.
\newblock GitHub repository, 2026{\natexlab{a}}.
\newblock URL \url{https://github.com/NVIDIA-NeMo/Gym}.

\bibitem[{Hugging Face}(2026)]{openenv}
{Hugging Face}.
\newblock {OpenEnv: A Standard Interface for Agent Environments}.
\newblock GitHub repository, 2026.
\newblock URL \url{https://github.com/huggingface/openenv}.

\bibitem[{HUD}(2026)]{hud}
{HUD}.
\newblock {HUD: Environments and Evaluation for Computer-Use Agents}.
\newblock Project website, 2026.
\newblock URL \url{https://hud.ai}.

\bibitem[{Strands Agents}(2026)]{strands}
{Strands Agents}.
\newblock {Strands Agents}.
\newblock Project website, 2026.
\newblock URL \url{https://strandsagents.com}.

\bibitem[{Sierra Research}(2024)]{taubench}
{Sierra Research}.
\newblock {$\tau$-bench: A Benchmark for Tool-Agent-User Interaction}.
\newblock GitHub repository, 2024.
\newblock URL \url{https://github.com/sierra-research/tau-bench}.

\bibitem[{Prime Intellect}(2026)]{verifiers}
{Prime Intellect}.
\newblock {Verifiers: Environments and Rubrics for LLM Reinforcement Learning}.
\newblock GitHub repository, 2026.
\newblock URL \url{https://github.com/PrimeIntellect-ai/verifiers}.

\bibitem[{KVCache.AI}(2026)]{agentenv}
{KVCache.AI}.
\newblock {AgentENV: Sandboxed Environments for Agent Training}.
\newblock GitHub repository, 2026.
\newblock URL \url{https://github.com/kvcache-ai/AgentENV}.

\bibitem[{Daytona}(2026)]{daytona}
{Daytona}.
\newblock {Daytona: Secure Infrastructure for Running AI-Generated Code}.
\newblock Project website, 2026.
\newblock URL \url{https://www.daytona.io}.

\bibitem[{E2B}(2026)]{e2b}
{E2B}.
\newblock {E2B: Sandboxes for AI Agents}.
\newblock Project website, 2026.
\newblock URL \url{https://e2b.dev}.

\bibitem[{Modal Labs}(2026)]{modal}
{Modal Labs}.
\newblock {Modal: Serverless Compute for AI Workloads}.
\newblock Project website, 2026.
\newblock URL \url{https://modal.com}.

\bibitem[{Miles Team}(2026{\natexlab{a}})]{miles_tito}
{Miles Team}.
\newblock {No Token Left Behind: Demystifying Token-In-Token-Out in Miles}.
\newblock LMSYS Org blog, May 2026{\natexlab{a}}.
\newblock URL \url{https://www.lmsys.org/blog/2026-05-13-no-token-left-behind}.

\bibitem[Ma et~al.(2025)Ma, Zhang, Zhao, Song, Wang, Sui, and Luo]{r3}
Wenhan Ma, Hailin Zhang, Liang Zhao, Yifan Song, Yudong Wang, Zhifang Sui, and
  Fuli Luo.
\newblock {Stabilizing MoE Reinforcement Learning by Aligning Training and
  Inference Routers}.
\newblock \emph{arXiv preprint arXiv:2510.11370}, 2025.
\newblock Introduces Rollout Routing Replay (R3).

\bibitem[{Miles Team}(2025)]{miles_fp8}
{Miles Team}.
\newblock {Unified FP8: Moving Beyond Mixed Precision for Stable and
  Accelerated MoE RL}.
\newblock LMSYS Org blog, November 2025.
\newblock URL \url{https://www.lmsys.org/blog/2025-11-25-fp8-rl}.

\bibitem[{Miles Team}(2026{\natexlab{b}})]{miles_int4}
{Miles Team}.
\newblock {Squeezing 1TB Model Rollout into a Single H200: INT4 QAT RL
  End-to-End Practice}.
\newblock LMSYS Org blog, January 2026{\natexlab{b}}.
\newblock URL \url{https://www.lmsys.org/blog/2026-01-26-int4-qat}.

\bibitem[{Miles Team}(2026{\natexlab{c}})]{miles_mxfp8_nvfp4}
{Miles Team}.
\newblock {Towards Blackwell-Native 8-bit and 4-bit RL: End-to-End MXFP8 and
  NVFP4 RL in Miles}.
\newblock LMSYS Org blog, July 2026{\natexlab{c}}.
\newblock URL \url{https://www.lmsys.org/blog/2026-07-29-mxfp8-nvfp4-rl}.

\bibitem[Cook et~al.(2025)Cook, Guo, Xiao, Lin, Wyss, Nazemi, Mishra, del
  Mundo, Blankevoort, and Han]{fourover6}
Jack Cook, Junxian Guo, Guangxuan Xiao, Yujun Lin, Keith Wyss, Mahdi Nazemi,
  Asit Mishra, Carlo del Mundo, Tijmen Blankevoort, and Song Han.
\newblock {Four Over Six: More Accurate NVFP4 Quantization with Adaptive Block
  Scaling}.
\newblock \emph{arXiv preprint arXiv:2512.02010}, 2025.

\bibitem[{NVIDIA}(2026{\natexlab{b}})]{megatron_bridge}
{NVIDIA}.
\newblock {Megatron Bridge: A Conversion and Verification Layer between Hugging
  Face and Megatron Core}.
\newblock GitHub repository, 2026{\natexlab{b}}.
\newblock URL \url{https://github.com/NVIDIA-NeMo/Megatron-Bridge}.

\bibitem[Zheng et~al.(2025{\natexlab{a}})Zheng, Liu, Li, Chen, Yu, Gao, Dang,
  Liu, Men, Yang, Zhou, and Lin]{gspo}
Chujie Zheng, Shixuan Liu, Mingze Li, Xiong-Hui Chen, Bowen Yu, Chang Gao, Kai
  Dang, Yuqiong Liu, Rui Men, An~Yang, Jingren Zhou, and Junyang Lin.
\newblock {Group Sequence Policy Optimization}.
\newblock \emph{arXiv preprint arXiv:2507.18071}, 2025{\natexlab{a}}.

\bibitem[Hu et~al.(2025)Hu, Liu, Xu, and Shen]{reinforcepp}
Jian Hu, Jason~Klein Liu, Haotian Xu, and Wei Shen.
\newblock {REINFORCE++: Stabilizing Critic-Free Policy Optimization with Global
  Advantage Normalization}.
\newblock \emph{arXiv preprint arXiv:2501.03262}, 2025.

\bibitem[Schulman et~al.(2017)Schulman, Wolski, Dhariwal, Radford, and
  Klimov]{ppo}
John Schulman, Filip Wolski, Prafulla Dhariwal, Alec Radford, and Oleg Klimov.
\newblock {Proximal Policy Optimization Algorithms}.
\newblock \emph{arXiv preprint arXiv:1707.06347}, 2017.

\bibitem[{Kimi Team}(2025)]{kimik2}
{Kimi Team}.
\newblock {Kimi K2: Open Agentic Intelligence}.
\newblock \emph{arXiv preprint arXiv:2507.20534}, 2025.

\bibitem[{Miles Team}(2026{\natexlab{d}})]{miles_p2p}
{Miles Team}.
\newblock {Updating 1T Parameters in Seconds --- P2P Weight Transfer in Large
  Scale Distributed RL}.
\newblock LMSYS Org blog, April 2026{\natexlab{d}}.
\newblock URL \url{https://www.lmsys.org/blog/2026-04-29-p2p-update}.

\bibitem[{Qwen Team}(2025{\natexlab{a}})]{qwen3}
{Qwen Team}.
\newblock {Qwen3 Technical Report}.
\newblock \emph{arXiv preprint arXiv:2505.09388}, 2025{\natexlab{a}}.

\bibitem[{GLM-5 Team}(2026)]{glm5}
{GLM-5 Team}.
\newblock {GLM-5: from Vibe Coding to Agentic Engineering}.
\newblock \emph{arXiv preprint arXiv:2602.15763}, 2026.

\bibitem[{Qwen Team}(2024)]{qwen25}
{Qwen Team}.
\newblock {Qwen2.5 Technical Report}.
\newblock \emph{arXiv preprint arXiv:2412.15115}, 2024.

\bibitem[{OpenAI}(2025)]{gptoss}
{OpenAI}.
\newblock {gpt-oss-120b \& gpt-oss-20b Model Card}.
\newblock \emph{arXiv preprint arXiv:2508.10925}, 2025.

\bibitem[{Kimi Team}(2026)]{kimik25}
{Kimi Team}.
\newblock {Kimi K2.5: Visual Agentic Intelligence}.
\newblock \emph{arXiv preprint arXiv:2602.02276}, 2026.

\bibitem[{Z.ai}(2026{\natexlab{a}})]{glm51_blog}
{Z.ai}.
\newblock {GLM-5.1}.
\newblock Z.ai blog, April 2026{\natexlab{a}}.
\newblock URL \url{https://z.ai/blog/glm-5.1}.

\bibitem[{Z.ai}(2026{\natexlab{b}})]{glm52_blog}
{Z.ai}.
\newblock {GLM-5.2}.
\newblock Z.ai blog, June 2026{\natexlab{b}}.
\newblock URL \url{https://z.ai/blog/glm-5.2}.

\bibitem[{Qwen Team}(2026{\natexlab{a}})]{qwen35}
{Qwen Team}.
\newblock {Qwen3.5: Towards Native Multimodal Agents}.
\newblock Qwen blog, February 2026{\natexlab{a}}.
\newblock URL \url{https://qwen.ai/blog?id=qwen3.5}.

\bibitem[{Qwen Team}(2026{\natexlab{b}})]{qwen36}
{Qwen Team}.
\newblock {Qwen3.6-35B-A3B: Agentic Coding Power, Now Open to All}.
\newblock Qwen blog, April 2026{\natexlab{b}}.
\newblock URL \url{https://qwen.ai/blog?id=qwen3.6-35b-a3b}.

\bibitem[{Thinking Machines Lab}(2026)]{inkling}
{Thinking Machines Lab}.
\newblock {Inkling: Our Open-Weights Model}.
\newblock Thinking Machines Lab news post, July 2026.
\newblock URL \url{https://thinkingmachines.ai/news/introducing-inkling/}.

\bibitem[Lu(2025)]{tml_opd}
Kevin Lu.
\newblock {On-Policy Distillation}.
\newblock Thinking Machines Lab: Connectionism, October 2025.
\newblock URL \url{https://thinkingmachines.ai/blog/on-policy-distillation/}.

\bibitem[Li et~al.(2026)Li, Zuo, He, Zhang, Xiao, Qian, Yu, Gao, Yang, Liu, and
  Ding]{rethinking_opd}
Yaxuan Li, Yuxin Zuo, Bingxiang He, Jinqian Zhang, Chaojun Xiao, Cheng Qian,
  Tianyu Yu, Huan-ang Gao, Wenkai Yang, Zhiyuan Liu, and Ning Ding.
\newblock {Rethinking On-Policy Distillation of Large Language Models:
  Phenomenology, Mechanism, and Recipe}.
\newblock \emph{arXiv preprint arXiv:2604.13016}, 2026.

\bibitem[{Miles Team}(2026{\natexlab{e}})]{miles_opd}
{Miles Team}.
\newblock {OPD Support in Miles}.
\newblock LMSYS Org blog, July 2026{\natexlab{e}}.
\newblock URL \url{https://www.lmsys.org/blog/2026-07-18-opd-support-in-miles}.

\bibitem[Yu et~al.(2025)Yu, Zhang, Zhu, Yuan, et~al.]{dapo}
Qiying Yu, Zheng Zhang, Ruofei Zhu, Yufeng Yuan, et~al.
\newblock {DAPO: An Open-Source LLM Reinforcement Learning System at Scale}.
\newblock \emph{arXiv preprint arXiv:2503.14476}, 2025.

\bibitem[Shah et~al.(2024)Shah, Bikshandi, Zhang, Thakkar, Ramani, and
  Dao]{flashattn3}
Jay Shah, Ganesh Bikshandi, Ying Zhang, Vijay Thakkar, Pradeep Ramani, and Tri
  Dao.
\newblock {FlashAttention-3: Fast and Accurate Attention with Asynchrony and
  Low-precision}.
\newblock \emph{arXiv preprint arXiv:2407.08608}, 2024.

\bibitem[He(2025)]{tml_nondeterminism}
Horace He.
\newblock {Defeating Nondeterminism in LLM Inference}.
\newblock Thinking Machines Lab: Connectionism, September 2025.
\newblock URL
  \url{https://thinkingmachines.ai/blog/defeating-nondeterminism-in-llm-inference/}.

\bibitem[Fang and Zhao(2024)]{usp}
Jiarui Fang and Shangchun Zhao.
\newblock {USP: A Unified Sequence Parallelism Approach for Long Context
  Generative AI}.
\newblock \emph{arXiv preprint arXiv:2405.07719}, 2024.

\bibitem[{Wan Team}(2025)]{wan}
{Wan Team}.
\newblock {Wan: Open and Advanced Large-Scale Video Generative Models}.
\newblock \emph{arXiv preprint arXiv:2503.20314}, 2025.

\bibitem[Liu et~al.(2025)Liu, Liu, Liang, Li, Liu, et~al.]{flowgrpo}
Jie Liu, Gongye Liu, Jiajun Liang, Yangguang Li, Jiaheng Liu, et~al.
\newblock {Flow-GRPO: Training Flow Matching Models via Online RL}.
\newblock \emph{arXiv preprint arXiv:2505.05470}, 2025.

\bibitem[Zheng et~al.(2025{\natexlab{b}})Zheng, Chen, Ye, Wang, Zhang, Jiang,
  Su, Ermon, Zhu, and Liu]{diffusionnft}
Kaiwen Zheng, Huayu Chen, Haotian Ye, Haoxiang Wang, Qinsheng Zhang, Kai Jiang,
  Hang Su, Stefano Ermon, Jun Zhu, and Ming-Yu Liu.
\newblock {DiffusionNFT: Online Diffusion Reinforcement with Forward Process}.
\newblock \emph{arXiv preprint arXiv:2509.16117}, 2025{\natexlab{b}}.

\bibitem[Esser et~al.(2024)Esser, Kulal, Blattmann, Entezari, et~al.]{sd3}
Patrick Esser, Sumith Kulal, Andreas Blattmann, Rahim Entezari, et~al.
\newblock {Scaling Rectified Flow Transformers for High-Resolution Image
  Synthesis}.
\newblock \emph{arXiv preprint arXiv:2403.03206}, 2024.

\bibitem[{Qwen Team}(2025{\natexlab{b}})]{qwenimage}
{Qwen Team}.
\newblock {Qwen-Image Technical Report}.
\newblock \emph{arXiv preprint arXiv:2508.02324}, 2025{\natexlab{b}}.

\bibitem[{Lightricks}(2026)]{ltx2}
{Lightricks}.
\newblock {LTX-2}.
\newblock GitHub repository, 2026.
\newblock URL \url{https://github.com/Lightricks/LTX-2}.

\bibitem[{NVIDIA}(2026{\natexlab{c}})]{cosmos3}
{NVIDIA}.
\newblock {Cosmos 3: Omnimodal World Models for Physical AI}.
\newblock \emph{arXiv preprint arXiv:2606.02800}, 2026{\natexlab{c}}.

\bibitem[{MiniMax}(2026)]{minimaxh3}
{MiniMax}.
\newblock {MiniMax H3: An Open Model Breaking the Boundaries Between Tasks and
  Modalities}.
\newblock MiniMax Research blog, July 2026.
\newblock URL \url{https://www.minimax.io/blog/minimax-h3}.

\bibitem[{SGLang Team} and {Miles Team}(2026{\natexlab{a}})]{day0_kimi_k3}
{SGLang Team} and {Miles Team}.
\newblock {SGLang and Miles Add Day-0 Support for Kimi K3}.
\newblock LMSYS Org blog, July 2026{\natexlab{a}}.
\newblock URL \url{https://www.lmsys.org/blog/2026-07-27-kimi-k3-day0-support}.

\bibitem[{SGLang Team} and {Miles Team}(2026{\natexlab{b}})]{day0_deepseek_v4}
{SGLang Team} and {Miles Team}.
\newblock {DeepSeek-V4 on Day 0: From Fast Inference to Verified RL with SGLang
  and Miles}.
\newblock LMSYS Org blog, April 2026{\natexlab{b}}.
\newblock URL \url{https://www.lmsys.org/blog/2026-04-25-deepseek-v4}.

\bibitem[{SGLang Team} and {Miles Team}(2026{\natexlab{c}})]{day0_qwen38}
{SGLang Team} and {Miles Team}.
\newblock {SGLang and Miles Add Day-0 Support for Qwen3.8}.
\newblock LMSYS Org blog, August 2026{\natexlab{c}}.
\newblock URL \url{https://www.lmsys.org/blog/2026-08-12-qwen3-8-day0-support}.

\bibitem[{SGLang Team} and {Miles Team}(2026{\natexlab{d}})]{day0_inkling}
{SGLang Team} and {Miles Team}.
\newblock {SGLang and Miles Add Day-0 Support for Inkling, a Frontier
  Multimodal Model}.
\newblock LMSYS Org blog, July 2026{\natexlab{d}}.
\newblock URL \url{https://www.lmsys.org/blog/2026-07-15-inkling-day0-support}.

\bibitem[{SGLang Team} and {Miles Team}(2026{\natexlab{e}})]{day0_nemotron3}
{SGLang Team} and {Miles Team}.
\newblock {SGLang and Miles Add Day-0 Support for NVIDIA Nemotron 3 Ultra for
  Long-Running Autonomous Agents}.
\newblock LMSYS Org blog, June 2026{\natexlab{e}}.
\newblock URL
  \url{https://www.lmsys.org/blog/2026-06-04-nvidia-run-nemotron-3-ultra}.

\bibitem[{DeepSeek-AI}(2025)]{deepseek_v32}
{DeepSeek-AI}.
\newblock {DeepSeek-V3.2: Pushing the Frontier of Open Large Language Models}.
\newblock \emph{arXiv preprint arXiv:2512.02556}, 2025.

\bibitem[{Moonshot AI}(2026)]{kimik26}
{Moonshot AI}.
\newblock {Kimi K2.6}.
\newblock Hugging Face model card, 2026.
\newblock URL \url{https://huggingface.co/moonshotai/Kimi-K2.6}.

\bibitem[{Gemma Team, Google DeepMind}(2026)]{gemma4}
{Gemma Team, Google DeepMind}.
\newblock {Gemma 4 Technical Report}.
\newblock \emph{arXiv preprint arXiv:2607.02770}, 2026.

\bibitem[{Miles Team}(2026{\natexlab{f}})]{miles_rocm}
{Miles Team}.
\newblock {ROCm Support for Miles: Large-Scale RL Post-Training on AMD Instinct
  GPUs}.
\newblock LMSYS Org blog, March 2026{\natexlab{f}}.
\newblock URL \url{https://www.lmsys.org/blog/2026-03-17-rocm-miles-rl-amd}.

\bibitem[{Miles Team}(2026{\natexlab{g}})]{miles_rocm_dsv4}
{Miles Team}.
\newblock {Bringing DeepSeek-V4 Flash RL Training to AMD Instinct MI355X GPUs
  with Miles}.
\newblock LMSYS Org blog, July 2026{\natexlab{g}}.
\newblock URL \url{https://www.lmsys.org/blog/2026-07-10-rocm-miles-dsv4}.

\bibitem[{Miles Team}(2026{\natexlab{h}})]{miles_repo}
{Miles Team}.
\newblock {Miles: A Full-Stack System for Frontier Post-Training}.
\newblock GitHub repository, 2026{\natexlab{h}}.
\newblock URL \url{https://github.com/radixark/miles}.

\end{thebibliography}

\end{document}